\documentclass[preprint,12pt]{elsarticle}

\usepackage{hyperref}
\usepackage{url}
\usepackage{tabularx}
\usepackage{enumitem}

\usepackage{amsmath}
\usepackage{amssymb}
\usepackage{mathtools}

\usepackage{amsthm}
\newtheorem{theorem}{Theorem}

\usepackage[dvipsnames]{xcolor}

\newcommand{\change}[1] {{\textcolor{Black}{#1}}}

\usepackage{multirow}

\usepackage{pgfplotstable}
\pgfplotsset{compat=1.18}
\usepackage{graphicx}
\usepackage{subcaption}

\usepackage{comment} 
\usepackage{booktabs}

\usepackage{physics} 

\usepackage{empheq}

\usepackage{amsmath}
\usepackage{amssymb}
\usepackage{bbm}
\usepackage{bm} 

\usepackage{algorithm}
\usepackage{algorithmic}

\usepackage{derivative}

\usepackage{cases}

\newcommand{\ucwt}{u_c^{\rm WT}}
\newcommand{\uctc}{u_c^{\rm TC}}

\newcommand{\bL}{\mathbf{L}}

\newcommand{\unn}{u_{\rm NN}}
\newcommand{\ufdm}{u_{\rm FDM}}
\newcommand{\ugt}{u_{\rm GT}}

\newcommand{\lres}{\mathcal{L}_{\rm res}}
\newcommand{\ldat}{\mathcal{L}_{\rm data}}
\newcommand{\lreg}{\mathcal{L}_{\rm reg}}
\newcommand{\llo}{\mathcal{L}_{\rm LO}} 

\newcommand{\lrespinn}{\mathcal{L}_{\rm res}^{\rm PINN}}
\newcommand{\ldatpinn}{\mathcal{L}_{\rm data}^{\rm PINN}}

\newcommand{\ldr}{\mathcal{L}_{\rm rgrad}}
\newcommand{\wdr}{w_{\rm rgrad}}

\newcommand{\ldatpre}{\mathcal{L}_{\rm u_0}}

\newcommand{\lopt}{\mathcal{L}_{\rm LO}}

\newcommand{\wdat}{w_{\rm data}}
\newcommand{\wreg}{w_{\rm reg}}

\newcommand{\tres}{\mathcal{T}_{\rm res}}

\newcommand{\tdat}{\mathcal{T}_{\rm data}}
\newcommand{\treg}{\mathcal{T}_{\rm reg}}

\newcommand{\cN}{\mathcal{N}}

\newcommand{\cF}{\mathcal{F}}
\newcommand{\cJ}{\mathcal{J}}

\newcommand{\wnn}{W}
\newcommand{\wfcn}{V}

\newcommand{\hu}{\hat{u}}
\newcommand{\bx}{\vb x}
\newcommand{\by}{\vb y}

\newcommand{\bD}{\vb{D}}

\newcommand{\bu}{\vb{u}}

\newcommand{\ubar}{\bar{u}}
\newcommand{\wbar}{\overline{W}}
\newcommand{\gapprox}{g_{\rm approx}}
\newcommand{\gtrue}{g_{\rm true}}

\begin{document}
\begin{frontmatter}

\title{BiLO: Bilevel Local Operator Learning for PDE Inverse Problems}

\author[uci_math]{Ray Zirui Zhang\corref{cor1}}
\ead{zirui.zhang@uci.edu}
\author[uci_math]{Christopher E. Miles}
\author[uci_cs]{Xiaohui Xie}
\author[uci_math,uci_bme]{John S. Lowengrub\corref{cor1}}
\ead{jlowengr@uci.edu}

\cortext[cor1]{Corresponding author}
\address[uci_math]{Department of Mathematics, University of California, Irvine}
\address[uci_cs]{Department of Computer Science, University of California, Irvine}
\address[uci_bme]{Department of Biomedical Engineering, University of California, Irvine}

\begin{abstract}
We propose a new neural network based method for solving inverse problems for partial differential equations (PDEs) by formulating the PDE inverse problem as a bilevel optimization problem. At the upper level, we minimize the data loss with respect to the PDE parameters. At the lower level, we train a neural network to locally approximate the PDE solution operator in the neighborhood of a given set of PDE parameters, which enables an accurate approximation of the descent direction for the upper level optimization problem.
The lower level loss function includes the least-square penalty of both the residual and its derivative with respect to the PDE parameters. 
We apply gradient descent simultaneously on both the upper and lower level optimization problems, leading to an effective and fast algorithm. The method, which we refer to as BiLO (Bilevel Local Operator learning), is also able to efficiently infer unknown functions in the PDEs through the introduction of an auxiliary variable. We provide a theoretical analysis that justifies our approach.
Through extensive experiments over multiple PDE systems, we demonstrate that our method enforces strong PDE constraints, is robust to sparse and noisy data, 
and eliminates the need to balance the residual and the data loss, which is inherent to the soft PDE constraints in many existing methods.
\end{abstract}

  
  \end{frontmatter}

\section{Introduction}


A fundamental task across various scientific and engineering fields is to infer the unknown parameters of a partial differential equation (PDE) from observed data. Applications include seismic imaging \cite{dengOpenFWILargeScaleMultiStructural2023,martinStochasticNewtonMCMC2012,yangSeismicWavePropagation2021}, electrical impedance tomography \cite{uhlmannElectricalImpedanceTomography2009,molinaroNeuralInverseOperators2023}, personalized medicine \cite{lipkovaPersonalizedRadiotherapyDesign2019,zhangPersonalizedPredictionsGlioblastoma2025,schaferBayesianPhysicsBasedModeling2021,subramanianEnsembleInversionBrain2023}, and climate modeling \cite{senGlobalOptimizationMethods2013}.
PDE inverse problems are commonly addressed within the frameworks of PDE-constrained optimization (PDECO) \cite{hinzeOptimizationPDEConstraints2008} or Bayesian inference \cite{stuartInverseProblemsBayesian2010}.
In the PDE constrained optimization framework, the objective is to minimize the difference between the observed data and the PDE solution, and the PDE is enforced as a constraint using adjoint or deep learning methods.
In the Bayesian inference framework, the inverse problem is formulated as a statistical inference problem, where the goal is to estimate the posterior distribution of the parameters given the data. This usually requires sampling parameter space and solving the forward PDE multiple times. 

This is the first paper in a two-part series. Here in Part I, we develop a constrained optimization framework for solving PDE inverse problems using deep learning. In Part II, we extend this approach to Bayesian inference frameworks \cite{zhangBiLOBilevelLocal2025}.




\subsection{Related work}

The \textbf{Adjoint Method} is widely used for computing the gradients of the objective function with respect to the PDE parameters using numerical PDE solvers in the PDE-constrained optimization framework.
This method provides accurate gradients and strong PDE constraints.
However, the adjoint method requires explicitly deriving the adjoint equation and solving both forward and adjoint equations at each iteration, which is complex and computationally expensive, especially for nonlinear or high-dimensional problems \cite{hinzeOptimizationPDEConstraints2008,plessixReviewAdjointstateMethod2006}.

\textbf{Physics-Informed Neural Networks (PINNs)} 
have emerged as novel methods for solving inverse problems in a PDE constrained optimization framework \cite{karniadakisPhysicsinformedMachineLearning2021,raissiPhysicsinformedNeuralNetworks2019,jagtapDeepLearningInverse2022,jagtapPhysicsinformedNeuralNetworks2022,chenPhysicsinformedNeuralNetworks2020,zhangPersonalizedPredictionsGlioblastoma2025,yangBPINNsBayesianPhysicsinformed2021,kapoorPhysicsinformedNeuralNetworks2024,chenPhysicsinformedNeuralNetworks2020,zhangPersonalizedPredictionsGlioblastoma2025,zongImprovedTrainingPhysicsinformed2023}.
PINNs represent PDE solutions using neural networks and embed both the data and the PDE into the loss function through a mesh-free approach.
By minimizing the total loss, PINNs effectively solve the PDE, fit the data, and infer the parameters simultaneously, showcasing integration of mathematical models with data-driven learning processes. 
A related approach, \textbf{Optimizing a Discrete Loss (ODIL)}, utilizes conventional numerical discretizations of the PDEs and the loss is minimized over the parameters and the PDE solutions at the grid points rather than the weights of a neural network
\cite{karnakovOptimizingDIscreteLoss2022,balcerakIndividualizingGliomaRadiotherapy2024}. 
However, in these methods, the PDE is enforced as a soft constraint, which requires balancing the residual and the data loss, and can lead to a trade-off between fitting the data and solving the PDE accurately.

\textbf{Neural Operators (NOs)} aim to approximate the PDE solution operator (parameter-to-solution map) and can serve as surrogate models for the forward PDE solvers \cite{kovachkiNeuralOperatorLearning2022,akyildizEfficientPriorCalibration2025,nelsenOperatorLearningMeets2025}. 
Once these surrogates are established, they can be integrated into a Bayesian inference framework or other optimization algorithms to solve inverse problems, leveraging the speed of evaluating a neural network \cite{zhouAIaidedGeometricDesign2024,pathakFourCastNetGlobalDatadriven2022,luMultifidelityDeepNeural2022,maoPPDONetDeepOperator2023}.
Some examples of operator learning frameworks include the Fourier Neural Operator (FNO) \cite{liFourierNeuralOperator2021,liPhysicsInformedNeuralOperator2024,whitePhysicsInformedNeuralOperators2023}, Deep Operator Network (DeepONet) \cite{luLearningNonlinearOperators2021,wangLearningSolutionOperator2021}, In-Context Operator (ICON) \cite{yangIncontextOperatorLearning2023}, among others, e.g. \cite{oleary-roseberryDerivativeInformedNeuralOperator2024,molinaroNeuralInverseOperators2023}. 
However, when used to solve inverse problems, neural operators can encounter challenges when the ground truth is out of the distribution of the training dataset.

There are many other methods for PDE inverse problems using deep learning; see \cite{nganyutanyuDeepLearningMethods2023,herrmannDeepLearningComputational2024,bruntonMachineLearningPartial2023} for more comprehensive reviews.

\subsection*{Main Contributions}
In this paper, we focus on solving PDE inverse problems in the PDE-constrained optimization framework using deep learning methods.
The contributions are as follows:
\begin{itemize}
  \item We formulate the PDE inverse problem as a bilevel optimization problem, where the upper level problem minimizes the data loss with respect to the PDE parameters, and the lower level problem involves training a neural network to approximate the PDE solution operator locally at given PDE parameters, enabling direct and accurate computation of the descent direction for the upper level optimization problem. 
  \item At the lower level problem, we introduce the ``residual-gradient'' loss, which is the least-square penalty of the derivative of the residual with respect to the PDE parameters. We show that this loss term compels the neural network to approximate the PDE solution for a small neighborhood of the PDE parameters, thus a ``local operator''.
  \item Extensive experiments over multiple PDE systems demonstrate that our novel formulation is both more accurate and more robust than other existing methods. It exhibits stronger PDE fidelity, robustness to sparse and noisy data, and eliminates the need to balance the residual and the data loss, a common issue in PDE-based soft constraints.
  \item We solve the bilevel optimization problem using gradient descent simultaneously on both the upper and lower level optimization problems, leading to an effective and fast algorithm. 
  \item We extend our method to infer unknown functions that are also parameterized by neural networks through an auxiliary variable. This bypasses the need to learn a high-dimensional local operator.
  \item We rigorously analyze the
difference between the exact gradient of the upper-level loss and the approximate gradient that results from inexact minimization of the lower level problem. We establish an error bound for the difference between the gradients that provides a theoretical foundation for our approach.
  
\end{itemize}
Our approach combines elements of PINNs, operator learning, and the adjoint method.
Our method is related to PINNs in that both use neural networks to represent the solution to the PDE, both use automatic differentiation to compute the PDE residual, and both aim to solve the PDE and infer the parameters simultaneously.  
However, in PINNs, the PDE-constraint is enforced as a regularization term (or soft constraint), leading to a trade-off between fitting the data and solving the PDE accurately.
Compared with operator learning, which solves the PDE for a wide range of parameters and requires a large amount of synthetic data for training,
our method only learns the operator local to the PDE parameters at each step of the optimization process and does not require a synthetic dataset for training.
Similar to the adjoint method, we aim to approximate the descent direction for the PDE parameters with respect to the data loss, but we do not require deriving and solving the adjoint equation.


The outline of this paper, Part I of our study on solving PDE inverse problems using deep learning methods, is as follows. In Section \ref{method}, we present and analyze the BiLO method and compare the formulation with other approaches (PINNs, NOs). In Section \ref{numerical experiments}, we apply the BiLO method to a collection of PDE inverse problems (elliptic, parabolic, hyperbolic) and compare the results to PINNs and NO. In Section \ref{conclusion}, we summarize our results. In the Appendices, we present details of the numerical analysis, numerical implementations, computational cost, sensitivity analyses (to hyperparameters), and additional numerical results.

\section{Method}
\label{method}
\subsection{Bilevel Local Operator Learning (BiLO) for PDE Inverse Problems}

In this section, we present BiLO for solving PDE-constrained optimization problems where we aim to infer the PDE parameters from observed data.
Let $u: \Omega \to \mathbb{R}$ be a function defined over a domain $\Omega \subset \mathbb{R}^d$,
and $\hu$ be the observed data, which might be noisy.
For time-dependent problems, we treat time $t$ as a special component of $\bx$, and $\Omega$ includes the temporal domain.
We consider the following PDE-constrained optimization problem:
\begin{equation}
  \begin{aligned}
    &\min_{\theta} \quad \lVert u -  \hu\lVert^2_2 \\
    &\textrm{s.t.} \quad \cF(D^ku(\bx),...,D u(\bx), u(\bx), \theta) = \mathbf{0}\\
  \end{aligned}
  \label{eq:opt_scalar}
\end{equation}
where $D^k$ is the $k$-th order derivative operator, $\theta$ represents the PDE parameters, and  $\cF$ denotes equality constraints that include the PDE, the boundary and initial conditions, and interface conditions, if needed, such as in elliptic interface problems.

For notational simplicity, we define the residual function of the PDE constraint to be
\begin{equation}
  r(\bx, \theta) := \cF(D^ku(\bx,\theta),...,D u(\bx,\theta), u(\bx,\theta), \theta)
\end{equation}
The dependence of $r$ on $u$ is implicit. The local operator $u$ is characterized by the following two conditions:
\begin{itemize}
  \item \textbf{Condition 1:} $r(\bx, \theta) = \mathbf{0}$,
  \item \textbf{Condition 2:} $\grad_{\theta} r(\bx, \theta) = \mathbf{0}$.
\end{itemize}
Condition 1 means that $u$ should solve the PDE at $\theta$. 
Condition 2 suggests that small variation of $\theta$ should lead to small variation of the residual. 
If the conditions are satisfied, then the derivative of the data loss with respect to $\theta$ will approximate the descent direction, 
and we can find the optimal $\theta$ by minimizing the data loss with respect to $\theta$ using a gradient descent algorithm.

\change{
To further clarify these conditions, we view \(\cF\) as a differential operator acting on the state $u$ and parameters $\theta$; the PDE is then expressed by the operator equation \(\cF[u,\theta]=\mathbf{0}\).
We defer details of underlying function space or regularity assumptions to the appendix when theoretical results are presented.
The ``global solution operator'', or the parameter-to-solution map, $\theta \to u(\cdot,\theta)$, satisfies $\cF[u(\cdot,\theta),\theta] = 0$ for all $\theta$.
Finding the global solution operator is generally challenging, and is unnecessary if we only want to find the descent direction to update the current $\theta$ in an optimization algorithm.
If we consider a local perturbation $\theta \to \theta+\delta\theta$, a formal Taylor expansion yields
\begin{equation}
\begin{aligned}
 \mathcal{F}[u(\bx,\theta+\delta\theta), \theta+\delta\theta]
= \mathcal{F}[u(\bx,\theta), \theta]
+ \left( \frac{\delta \mathcal{F}}{\delta u}\pdv{u}{\theta}
+ \pdv{\mathcal{F}}{\theta} \right)\delta\theta
+ O(\delta\theta^2),
\end{aligned}
\label{eq:variation}
\end{equation}
where $\delta \mathcal{F}/\delta u$ denotes the Fréchet derivative of $\mathcal{F}$ with respect to $u$.
Then condition 1 corresponds to setting the zero-th order term in the expansion to zero: 
$$
r(\bx, \theta) = \mathcal{F}[u(\bx,\theta), \theta]=0,
$$
while condition 2 corresponds to setting the first-order term to zero:
$$
\grad_{\theta} r(\bx, \theta) = \dv{ }{\theta} \mathcal{F}[u(\bx,\theta), \theta]
= \frac{\delta \mathcal{F}}{\delta u}\pdv{u}{\theta}
+ \pdv{\mathcal{F}}{\theta}=0.
$$
Thus, a local operator solves the PDE at $\theta$ and is a second-order accurate approximation in $\delta\theta$ to the PDE under 
small perturbations of $\theta$.
}

We approximate the operator locally at $\theta$ using a neural network $u(\bx, \theta; \wnn)$, where $\wnn$ denotes the weights of the network.
The objective function \eqref{eq:opt_scalar} leads to the data loss:
\begin{equation}
  \label{eq:op_dat_loss}
  \ldat(\theta, \wnn) = \frac{1}{|\tdat|} \sum_{\bx \in \tdat} \left| u( \bx, \theta; \wnn) - \hu(\bx) \right|^2,
\end{equation}
where $\tdat$ denotes the set of \change{measurement locations for observed data}.
The residual loss is given by
\begin{equation}
  \label{eq:new_res_scalar_loss}
  \lres(\wnn,\theta) := \frac{1}{|\tres|} \sum_{\bx \in \tres} \left| r( \bx, \theta; \wnn) \right|^2.
\end{equation}
where $\tres$ is the set of collocation points for evaluating the residual.
\change{We introduce the following ``residual-gradient loss''}
\begin{equation}
  \label{eq:rg_loss}
  \ldr (\theta, \wnn) = \frac{1}{|\tres|} \sum_{\bx \in \tres} \left| \grad_{\theta} r( \bx, \theta) \right|^2.
\end{equation}
We define the ``local operator loss'' as the sum of the residual loss and the residual-gradient loss with weight $\wdr$:
\begin{equation}
  \lopt(\theta, \wnn) = \lres(\theta, \wnn) + \wdr \ldr(\theta, \wnn)
\end{equation}

Finally, we propose to solve the following bilevel optimization problem:
\begin{equation}
  \label{eq:bi_scalar}
  \begin{cases}
    \theta^* = \arg\min_{\theta} \ldat(\theta, \wnn^*(\theta)) \\
    \wnn^*(\theta) = \arg\min_{\wnn} \lopt(\theta, \wnn) \\
  \end{cases}
\end{equation}
In the upper level problem, we find the optimal PDE parameters $\theta$ by minimizing the data loss with respect to $\theta$.
In the lower level problem, we train a network to approximate the local operator $u(\bx, \theta; \wnn)$ by minimizing the local operator loss with respect to the weights of the neural network.

Boundary and initial conditions can be incorporated as additional loss terms, evaluated at respective domain boundaries. 
In some cases, these conditions can be enforced exactly by transforming the network output or specialized network architecture \cite{dongMethodRepresentingPeriodic2021,luPhysicsInformedNeuralNetworks2021,sukumarExactImpositionBoundary2022}.
For example, on the domain $\Omega = [0,1]$, one can impose the Dirichlet boundary condition $u(0)=u(1)=0$ by multiplying the output of the neural network by $x(1-x)$.
For simplicity of discussion, we focus on the residual loss and the data loss, and assume that the boundary conditions are enforced exactly.

\begin{figure}[!h]
  \centering
  \includegraphics[width=0.7\linewidth,keepaspectratio]{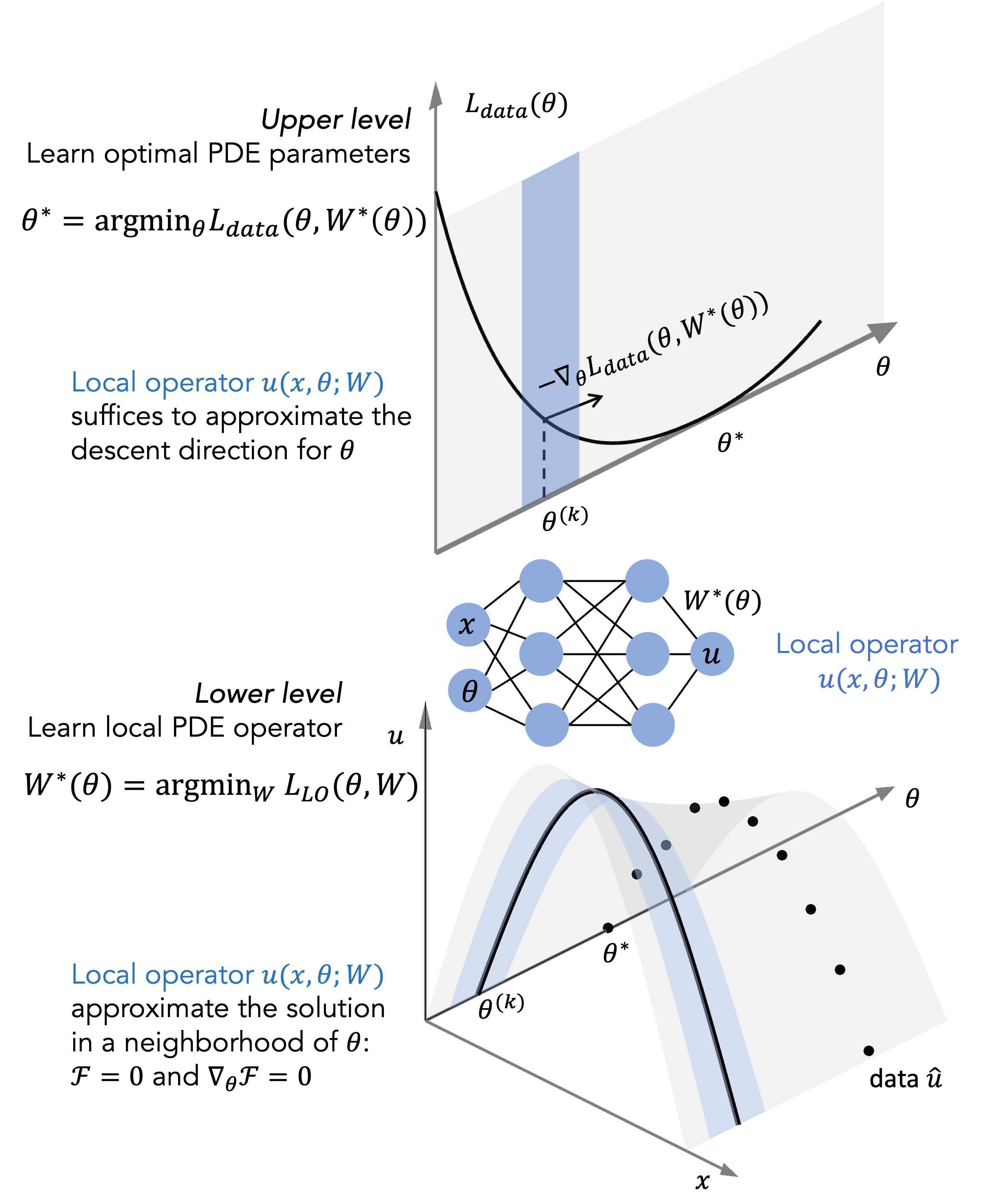}
  \caption{A schematic of BiLO. Top:
  The full PDE operator $u(\bx,\theta)$ (gray) solves the PDE for all $\theta$, while the local operator (blue) approximates the solution in a small neighborhood of $\theta$. The local operator is sufficient for approximating the descent direction of the data loss. The figure uses the model boundary value problem $-\theta u_{xx}=\sin(\pi x)$ with Dirichlet boundary condition.}
  \label{f:schema}
\end{figure}

Fig.~\ref{f:schema} illustrates the idea of the BiLO framework.
We consider functions of the form $u(\bx,\theta)$, where $\bx\in\Omega$ and $\theta\in\Theta$, where $\Theta$ is an admissible set of PDE parameters.
We call such a function the ``PDE solution operator'' (hereafter referred to as the ``operator''), if it solves the PDE for all $\theta$, that is, the map $\theta \mapsto u(\cdot,\theta)$ is the parameter-to-solution map.
Such an operator exists if the PDE solution is unique and continuous with respect to the parameters $\theta$.
An example of such an operator $u(\bx,\theta)$ is shown as a gray surface in Fig.~\ref{f:schema}, which solves the PDE $-\theta u_{xx}=\sin(\pi x)$ with Dirichlet boundary condition for all $\theta>0$. 
If such an operator is available, we can solve the optimization problem easily by minimizing the objective function using a gradient descent algorithm.
However, finding the full operator $u(\bx,\theta)$ is challenging and unnecessary.
Since we are only interested in the descent direction to update $\theta$, a local approximation of the solution operator suffices (blue surface in Fig.~\ref{f:schema}), that is,
the operator should approximate the PDE solution for a small neighborhood of a particular value of $\theta$. 


\subsection{Pre-training and Fine-tuning}

In this work, we assume access to an initial guess of the PDE parameters, $\theta_0$, as required by most gradient-based methods. In BiLO, the lower-level problem must be solved to compute the descent direction for the upper-level optimization. To this end, we introduce a pre-training phase in which we fix $\theta = \theta_0$ and train the neural network to approximate the local solution operator at $\theta_0$. Since $\theta$ is fixed, this stage is not a bilevel optimization problem, as only the lower-level problem is solved.
After pre-training, we solve the full bilevel optimization problem to infer the PDE parameters $\theta$, and we refer to this as the fine-tuning phase.

When available, a numerical solution of the PDE at $\theta_0$, denoted by $u_0(\bx)$ and computed using a method such as finite difference (FDM) or finite element (FEM), can be used to accelerate pre-training. We define a pre-training data loss $\ldatpre$ as the mean squared error between the numerical solution $u_0$ and the neural network output at $\theta_0$:
\begin{equation}
\ldatpre(\wnn) = \frac{1}{|\tres|} \sum_{\bx \in \tres} \left| u( \bx, \theta_0; \wnn) - u_0(\bx) \right|^2.
\end{equation}
The pre-training objective is then:
\begin{equation}
\min_{\wnn} \lopt(\theta_0, \wnn) + \ldatpre(\wnn).
\end{equation}
The use of $\ldatpre$ is optional, but can significantly speed up the pre-training process. This is computationally inexpensive, as we only need one numerical solution.
This strategy was been shown to be effective in \cite{zhangPersonalizedPredictionsGlioblastoma2025}, and is conceptually related to curriculum learning \cite{krishnapriyanCharacterizingPossibleFailure2021}, where the network first learns to approximate a simpler solution.


\subsection{Network Architecture}

\change{The network architecture follows the standard multilayer perceptron (MLP) design.
For the scalar parameter case, the embedding layer maps the inputs $\bx\in\mathbb{R}^d$ and the unknown PDE parameters $\theta\in\mathbb{R}^m$ to a high-dimensional vector $\mathbf{h}^{(0)}\in\mathbb{R}^p$, using an affine transformation followed by a non-linear activation function $\sigma$ ($\tanh$ in this work): 
\begin{equation}
  \mathbf{h}^{(0)} = \sigma(W^{(0)}_{\bx}\bx + W^{(0)}_{\theta} \theta + \mathbf{b}),
\end{equation}
where $W^{(0)}_{\bx} \in \mathbb{R}^{p \times d}$ is the embedding matrix for $\bx$, $W^{(0)}_{\theta} \in \mathbb{R}^{p \times m}$ is the embedding matrix for $\theta$, and $\mathbf{b}\in\mathbb{R}^p$ is the bias vector.
This transformation serves as the standard input embedding, mapping both coordinates and PDE parameters into a shared latent space and enables the network $u(\bx,\theta;\wnn)$ to model the nonlinear effect of $\theta$ on $u$ at each point $\bx$.
The embedding vector $\mathbf{h}^{(0)}$ is then passed through a series of fully connected layers with activation functions:
$\mathbf{h}^{(l)} = \sigma(W^{(l)} \mathbf{h}^{(l-1)} + b^{(l)})$ for $l=1,...,L-1$, where $L$ is the total number of layers, $W^{(l)} \in \mathbb{R}^{p \times p}$ and $b^{(l)}\in\mathbb{R}^p$ are the weights and bias of the $l$-th layer.
The final output is given by
$\mathbf{y} = W^{(L)} \mathbf{h}^{(L-1)} + b^{(L)}$, where $W^{(L)} \in \mathbb{R}^{1 \times p}$ and $b^{(L)}\in\mathbb{R}$.
Collectively, the network is denoted as $\cN( \bx,\theta; \wnn)$, where $\wnn = \{W^{(l)}, b^{(l)}\}_{l=0}^L$ denotes all the trainable weights of the neural network.
In some cases, a final transformation is applied to the output of the neural network
$u(\bx; \wnn) = \tau \left( \cN( \bx, \theta; \wnn), \bx \right)$,
to enforce  boundary conditions \cite{dongMethodRepresentingPeriodic2021,luPhysicsInformedNeuralNetworks2021,sukumarExactImpositionBoundary2022}.
For example, if the PDE is defined on a unit interval \([0, 1]\) with Dirichlet boundary conditions \(u(0) = u(1) = 0\), the BiLO solution can be represented as
$u(x, \theta; \wnn) = \cN(x,\theta;W) (1 - x) x$.
This MLP architecture is presented as a basic example, and more advanced architectures can also be incorporated, including residual connections \cite{heDeepResidualLearning2015}, random Fourier features \cite{wangEigenvectorBiasFourier2021}, and other modified MLP designs \cite{wangUnderstandingMitigatingGradient2021}.
}

\subsection{Inferring an unknown function}
\label{ss:learnfun}

We can extend our method to learn an unknown function $f(\bx)$ in the PDE,
such as a variable diffusion coefficient in the Poisson equation or an initial condition in the heat equation.
In these cases, the following PDE constrained optimization problem is solved:
\begin{equation}
  \begin{aligned}
    &\min_{f} \quad \lVert u -  \hu\lVert^2 + \wreg \lVert\grad f \rVert^2 \\
    &\textrm{s.t.} \quad \cF(D^ku(\bx),...,D u(\bx), u(\bx), f(\bx)) = \mathbf{0}\\
  \end{aligned}
\end{equation}
where the PDE depends on an unknown function $f$.
Given that these problems are ill-posed, regularization of the unknown function is often necessary.
A typical choice is the L2-norm of the gradient of the unknown function, which penalizes non-smooth functions.
The choice of an appropriate regularization form is important and problem-dependent. This paper assumes such choices are predetermined and are not  aspects of the method under direct consideration.

Suppose $f$ is parameterized by a neural network $f(\bx;\wfcn)$ with weights $\wfcn$.
A straightforward extension from the scalar parameter case is to learn the local operator of the form $u(\bx,\wfcn)$.
However, this would be computationally expensive, as the weights $\wfcn$ can be very high dimensional.
We introduce an auxiliary variable $z = f(\bx)$,
and find a local operator $u(\bx, z)$ such that $u(\bx, f(\bx))$ solves the PDE locally at $f$.
We define the following function $a$, which is the augmented residual function with an auxiliary variable $z$:
\begin{equation}
  a(\bx, z) := \cF(D^ku(\bx,z),...,D u(\bx,z), u(\bx,z), z)
\end{equation}
where $D = \nabla_{\bx} + \nabla_{\bx} z  \nabla_{z}$ is the total derivative operator.
If $u$ is a local solution operator at $f$, then we should have: (1) $a(\bx, f(\bx)) = 0$ so that the function $u(\bx, f(\bx))$ have zero residual, and
(2) $\grad_{z} a(\bx, f(\bx)) = 0$ so that small variations of $f$ should lead to small variations in the residual, as in the scalar parameter case in Eq. \eqref{eq:rg_loss}. These two conditions translate into the corresponding residual loss and residual-gradient loss, similar to Eqs.\eqref{eq:new_res_scalar_loss} and \eqref{eq:rg_loss}. 
The residual loss is given by
\begin{equation}
  \label{eq:new_res_fun_loss}
  \lres(\wnn,\wfcn) := \frac{1}{|\tres|} \sum_{\bx \in \tres} \left| a( \bx, f(\bx;\wfcn); \wnn) \right|^2.
\end{equation}
and the residual-gradient loss is 
\begin{equation}
  \label{eq:new_rg_fun_loss}
  \ldr (\wnn,\wfcn) = \frac{1}{|\tres|} \sum_{\bx \in \tres} \left|  \nabla_{z} a(\bx, f(\bx;\wfcn); \wnn)  \right|^2
\end{equation}
The data loss is similar to the parameter inference case in Eq. \eqref{eq:op_dat_loss} and depends on both $\wfcn$ and $\wnn$.
We also need the regularization loss, evaluated on $\treg$:
\begin{equation}
  \lreg(\wfcn) = \frac{1}{|\treg|}\sum_{\bx \in \treg} |\grad_{\bx} f(\bx;\wfcn)|^2.
\end{equation}

Finally, we solve the following bilevel optimization problem:
\begin{numcases}{}
  \wfcn^* = \arg \min_{\wfcn} \ldat(\wnn^*(\wfcn),  \wfcn) + \wreg \lreg(\wfcn) \label{eq:minpde}
  \\
  \wnn^*(\wfcn) = \arg \min_{\wnn} \lopt(\wnn, \wfcn) \label{eq:minnn}
\end{numcases}
where $\lopt = \lres + \wdr \ldr$.
At the upper level, we minimize the data loss and the regularization loss with respect to the weights $\wfcn$ of the unknown function, and at the lower level, we minimize the local operator loss with respect to the weights $\wnn$ of the local operator.
The pre-training stage is similar to the parameter inference case.
Given an initial guess of the unknown function $f_0$, and its corresponding numerical solution $u_0$, we can train the network $f_{\wfcn}$ to approximate $f_0$ by minimizing the MSE between $f_{\wfcn}$ and $f_0$,
and train the network $u_{\wnn}$ to be the local operator at $f_0$ by minimizing the local operator loss and the MSE between $u_{\wnn}$ and $u_0$.

\subsection{Algorithm and Theoretical Analysis}
\label{ss:alg}
\paragraph{Algorithm}
Solving bilevel optimization problems remains an active area of research \cite{zhangIntroductionBilevelOptimization2023,khanduriLinearlyConstrainedBilevel2023,yeBOMEBilevelOptimization2022,shenPenaltybasedBilevelGradient2023,shabanTruncatedBackpropagationBilevel2019,hongTwoTimescaleFrameworkBilevel2022,nelsenBilevelOptimizationLearning2025}.
In our case, the upper level problem is usually non-convex, and the lower level problem has a challenging loss landscape \cite{krishnapriyanCharacterizingPossibleFailure2021,basirCriticalInvestigationFailure2022}.
However, the lower level problem does not need to be solved to optimality at each iteration because the primary goal is to approximate the descent direction for the upper level problem.
We propose to apply gradient descent to the upper and lower level optimization problems simultaneously.
In Algorithm.~\ref{alg}, we describe the optimization algorithm for inferring scalar parameters in the BiLO framework.
The algorithm for inferring unknown functions is similar.
We write the algorithm as simple gradient descent for notational simplicity, while in practice we use ADAM \cite{kingmaAdamMethodStochastic2017} or its variants.
We can have two different learning rates for the two groups of variables $\wnn$ and $\theta$, denoted as $\alpha_{\wnn}$ and $\alpha_{\theta}$, respectively. 

\begin{algorithm}
  \caption{Bi-level Local Operator for inferring scalar PDE parameters}
  \begin{algorithmic}[1] 
    \STATE \textbf{Input:}  Collections of collocation points $\tres$ and $\tdat$, initial guess of the PDE parameters $\theta_0$
    \STATE \textbf{Pre-train:} Solve the lower level problem at fixed $\theta_0$:
      \begin{equation}
        \min_{\wnn} \lopt(\theta_0, \wnn)
      \end{equation}
    \STATE \textbf{Fine-Tune:} Simultaneous gradient descent at the upper and lower levels in system \eqref{eq:bi_scalar}.
    \begin{equation}
      \label{eq:bilogd}
      \begin{cases}
        \theta^{(k+1)} = \theta^{(k)} - \alpha_{\theta} \grad_{\theta} \ldat(\theta^{(k)}, \wnn^{(k)})\\
        \wnn^{(k+1)} = \wnn^{(k)} - \alpha_{\wnn} \grad_{\wnn} \lopt(\theta^{(k)}, \wnn^{(k)})
      \end{cases}
      \end{equation}
  \end{algorithmic}
  \label{alg}
\end{algorithm}

\paragraph{Theoretical Analysis}

 We next provide a theoretical characterization of our bilevel optimization method by analyzing the difference between the exact and approximate gradients of the upper-level loss. The approximate gradient arises from inexact minimization of the lower level problem. The exact gradient, or hypergradient, accounts for the total dependence of the system on the hyperparameter $\theta$, including the sensitivity of the ideal weights $W^*(\theta)$. In contrast, our simultaneous training algorithm uses an approximate gradient, which efficiently computes only the partial derivative with respect to $\theta$ at the current weights $\wbar$. Our analysis establishes two key results:

\begin{itemize}[leftmargin=*, itemsep=0pt]
  \item Consistency: We demonstrate that under ideal conditions (i.e., the lower-level problem is solved exactly), the approximate gradient is identical to the true gradient. A precise statement of the corresponding theorem (Theorem \ref{thm:consistency}) is given in \ref{ap:consistency} along with its proof.
  \item Approximation Error: More practically, we establish an error bound. Theorem \ref{thm:hypergrad_err_bound} (stated precisely and proved in \ref{ap:approximation}) guarantees that when the lower-level problem is solved to a tolerance $\epsilon$, the error between the approximate and true gradients is also bounded by $\epsilon$, assuming the PDE is well-behaved.
\end{itemize}

These theorems provide a solid theoretical foundation for our approach. Furthermore, our numerical experiments demonstrate the method's effectiveness under even less restrictive conditions than required by the theory.

\subsection{Difference between BiLO, PINN, and NO}

We next clarify the differences between BiLO, PINNs, and neural operators.

\textbf{Neural Operators} can serve as surrogate models for PDE solution operators, and can be used in algorithms that require solving the forward PDE multiple times, such as Bayesian inference, derivative-free optimization \cite{kaltenbachSemisupervisedInvertibleNeural2023,luMultifidelityDeepNeural2022}, and gradient-based optimization algorithms \cite{zhouAIaidedGeometricDesign2024,luMultifidelityDeepNeural2022,yangRapidSeismicWaveform2023}.
However, if the objective is to estimate parameters from limited data, the considerable initial cost for data generation and network training might seem excessive. 
The accuracy of specific PDE solutions depends on the accuracy of the neural operator, 
which may decrease if the true PDE parameters fall outside the training data's distribution \cite{dehoopCostAccuracyTradeOffOperator2022}.
This issue can be mitigated by instance-wise fine-tuning using the residual loss \cite{liPhysicsInformedNeuralOperator2024,wangLearningSolutionOperator2021},
though it introduces an additional trade-off: fine-tuning for one parameter set could reduce the operator's overall accuracy for other parameters and
an ``anchor loss'' is thus required to maintain generalization \cite{liPhysicsInformedNeuralOperator2024}.
Thus, in the context of finding the best estimate of the parameters given the data in a PDE-constrained optimization framework, we mainly compare BiLO with PINNs.

Within the \textbf{PINN} framework, the solution of the PDE is represented by a deep neural network $u(\bx; \wnn)$ \cite{karniadakisPhysicsinformedMachineLearning2021,raissiPhysicsinformedNeuralNetworks2019,luDeepXDEDeepLearning2021}.
Notice that the PDE parameters $\theta$ are not input to the neural network.
Therefore, the data loss does not depend on the PDE parameters $\theta$ directly, and we write the data loss as $\ldatpinn$.
$$\ldatpinn(W) =  \sum_{\bx \in \tdat}(u(\bx;W) - \hat{u}(\bx))^2$$
and enforce the PDE constraints by minimizing the residual loss.
$$\lrespinn(W,\theta) = \sum_{\bx \in \tres} \cF(D^ku(\bx;W),...,u(\bx;W),\theta)^2.$$

Solving an inverse problem using PINN involves minimizing an unconstrained optimization problem, where the objective function is the weighted sum of the residual loss and the data loss
\begin{equation}
  \label{eq:pinn_total}
  \min_{\wnn,\theta} \lrespinn(\wnn, \theta) + \wdat \ldatpinn(\wnn)
\end{equation}
where $\wdat$ is the weight of the data loss.
For simplicity of discussion, we assume the weight of the residual loss is always 1.
In PINN, the PDE is enforced as a soft constraint or as a regularization term for fitting the data. 
The relationship between the PDE parameter and the data loss is indirect. 
If we consider the gradient descent dynamics for training of the PINN, we have
\begin{equation}
  \label{eq:pinngd}
  \begin{cases}
    \theta^{k+1} = \theta^{k} - \alpha_{\theta} \grad_{\theta} \lrespinn(\wnn^{k}, \theta^{k})\\
    \wnn^{k+1} = \wnn^{k} - \alpha_{\wnn} \grad_{\wnn} (\lrespinn(\wnn^{k}, \theta^{k}) + \wdat \ldatpinn(\wnn^{k}))
  \end{cases}
\end{equation}
The descent directions of the PDE parameters do not directly depend on the data loss $\ldatpinn$.



\paragraph{Challenges for PINNs}
Solving PDE inverse problems using PINNs can encounter challenges stemming from the soft PDE constraint in Eq. \eqref{eq:pinn_total}, especially when the data is sparse and noisy, or when the PDE model does not fully explain the data \cite{zhangPersonalizedPredictionsGlioblastoma2025}.
The soft PDE constraint can result in a trade-off between fitting the data and solving the PDE accurately.
In addition, since the PDE parameters are updated in the descent direction of the residual loss, they can be biased toward parameters corresponding to very smooth solutions. 
Nevertheless, it is important to recognize that PINNs can indeed be effective for PDE inverse problems, if the weights are chosen properly \cite{kapoorPhysicsinformedNeuralNetworks2024,chenPhysicsinformedNeuralNetworks2020,jagtapPhysicsinformedNeuralNetworks2022}.


There are many techniques to improve the performance of PINNs,
such as adaptive sampling and weighting of collocation points \cite{nabianEfficientTrainingPhysicsinformed2021,wuComprehensiveStudyNonadaptive2023,luDeepXDEDeepLearning2021,anagnostopoulosResidualbasedAttentionPhysicsinformed2024}, new architectures \cite{jagtapAdaptiveActivationFunctions2020,wangPirateNetsPhysicsinformedDeep2024,wangUnderstandingMitigatingGradient2021,moseleyFiniteBasisPhysicsinformed2023},
new optimization algorithms \cite{basirPhysicsEqualityConstrained2022,krishnapriyanCharacterizingPossibleFailure2021}, new loss functions \cite{wang$L^2$PhysicsInformed2022,yuGradientenhancedPhysicsinformedNeural2022,sonSobolevTrainingPhysics2021}, adaptive weighting of loss terms \cite{madduInverseDirichletWeighting2022,wangUnderstandingMitigatingGradient2021,mcclennySelfAdaptivePhysicsInformedNeural2022,wangExpertsGuideTraining2023}.
However, these techniques do not fundamentally change the soft PDE-constraints in the PINN framework.
In our work, we propose a different optimization problem that does not involve a trade-off between the residual loss and the data loss,
and our method can be used in conjunction with many of these techniques to improve the performance.
Therefore, in the following numerical experiments, we do not use any of these techniques, and we focus on comparing the two different optimization formulations (BiLO and the soft PDE-constraints).

\change{We note that while our lower-level problem also involve a weighted sum of two loss terms (the residual loss and the residual-gradient loss), the
residual-gradient term does not compete with the residual loss in the way that the data loss does in PINNs.
The minimizer of the data term is an interpolant of the observations, whereas the minimizer of the residual term is a function that satisfies the PDE; these optima typically do not coincide, creating an inherent tension between fitting data and solving the PDE in PINNs.
In contrast, a local operator is a function characterized by vanishing residual and residual-gradient.
Thus, minimizing these two terms does not pull the neural network in conflicting directions. In the ideal case with sufficient network capacity, both terms can be sufficiently minimized simultaneously.
Therefore, the parameter $w_{\rm rgrad}$ conditions the optimization landscape rather than altering the location of the minimizer. We illustrate its effect in the Appendix.}

The challenge of balancing trade-offs also motivated the Bilevel PINN (BPN) method developed in \cite{haoBilevelPhysicsInformedNeural2023}, which applies a bilevel optimization framework to PDE inverse problems by representing the PDE solution with a neural network, using the residual loss for the lower-level problem, and approximating the upper-level hypergradient with Broyden's method. In contrast, our approach incorporates the PDE parameter as part of the network input, with the lower-level problem focused on approximating the local operator, allowing more direct computation of the upper-level descent direction. We compare BPN and BiLO in \ref{ap:bpn}.

\section{Numerical Experiments}
\label{numerical experiments}
We evaluate the effectiveness of our method on a diverse set of PDE inverse problems, encompassing elliptic, parabolic, and hyperbolic systems. Our test cases include challenging scenarios such as nonlinear dynamics, singular forcing terms, and inverse problems in Glioblastoma using patient data.
In our experiments, we denote the neural network solution (obtained via BiLO, PINN, or NO) as $\unn$, and the numerical solution computed with the estimated parameters using the Finite Difference Method (FDM) as $\ufdm$, which is computed to high accuracy and serves as the exact solution to measure the accuracy of the neural network solution.
To add synthetic noise, we consider Gaussian noise with mean 0 and standard deviation $\sigma$.

For each numerical experiment, we solve the optimization problem 5 times with different random seeds,
which affect both the initialization of the neural network and the noise (if applicable).
Although each realization of the noise may yield a different globally optimal PDE parameter $\theta^*$, the average of the estimated parameters across multiple runs should still be close to the ground truth parameter $\theta_{GT}$.
Therefore, we report the mean and standard deviation of the error between the estimated quantities and the ground truth.
We empirically determined $\wdr = 0.1$ and $\alpha_{\wnn} = \alpha_{\theta} = 0.001$ to be effective across our numerical experiments.

\subsection{Fisher-KPP Equation}

We aim to infer the unknown parameters $D$ and $\rho$ in the following nonlinear reaction-diffusion equation (Fisher-KPP equation) as in \cite{zouCorrectingModelMisspecification2024}:
\begin{equation}
  \begin{cases}
    u_t(x,t) = 0.01 D u_{xx}(x,t) + \rho u(1-u)\\
    u(x,0) = \frac{1}{2}\sin(\pi x)^2,\\ 
    u(0,t) = u(1,t) = 0.  \\
  \end{cases}
\end{equation}
The initial guesses of the PDE parameters are $D_0 = 1$ and $\rho_0 = 1$, and the ground truth parameters are $D_{GT} = 2$ and $\rho_{GT} = 2$.
\change{Thus the relative error of the initial guesses are $50\%$.}
This equation has been used to model various biological phenomena, such as
the growth of tumors \cite{swansonQuantitativeModelDifferential2000,harpoldEvolutionMathematicalModeling2007}
or the spreading of misfolded proteins \cite{schaferNetworkDiffusionModeling2020,schaferBayesianPhysicsBasedModeling2021, zhangDiscoveringReactionDiffusion2024}.

\subsubsection{Visualizing BiLO}\label{ss:fk}

\begin{figure}[!h]
\centering
\includegraphics[width=0.7\linewidth,keepaspectratio]{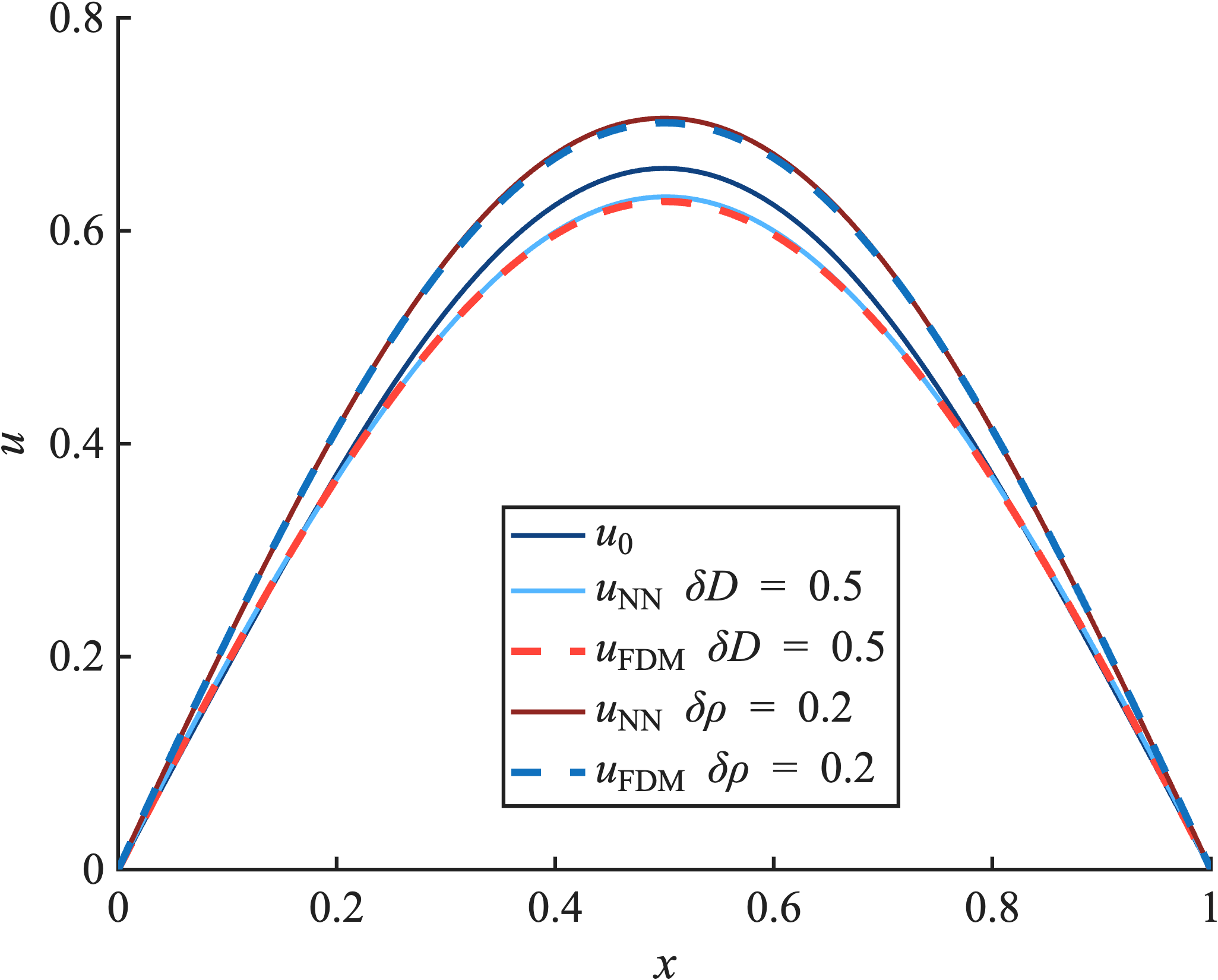}
\caption{
  Evaluating the local operator $u(\bx,D_0+\delta D,\rho_0 + \delta \rho)$ at $t=1$ after pretraining at $D_0$ and $\rho_0$ provides an approximation to the corresponding FDM solutions.}
  \label{f:fkvar}
\end{figure}

In Fig.~\ref{f:fkvar}, we visualize the local operator $u(x, D, \rho;\wnn)$ after pre-training with $D_0 = 1$ and $\rho_0 = 1$.
We consider the variation 
$(\delta D, \delta \rho)$ = $(0.5,0)$ and $(0,0.2)$ 
and evaluate the neural network at $u(x, D_0 + \delta D, \rho_0 + \delta \rho)$.
The FDM solutions of the PDE corresponding to the neighboring parameters are also shown.
The neural network approximates the solution corresponding to the neighboring parameters well, and the neural network is able to learn the local operator of the PDE. 
\change{
  Note that this figure is intended to show that the neural network varies consistently with the PDE constraint. In practice, we do not evaluate the network at nearby parameter values.
  Rather, Conditions 1 and 2 provide the correct gradient direction for updating the parameters.
}

\begin{figure}[!h]
  \centering
  \includegraphics[width=0.7\linewidth,keepaspectratio]{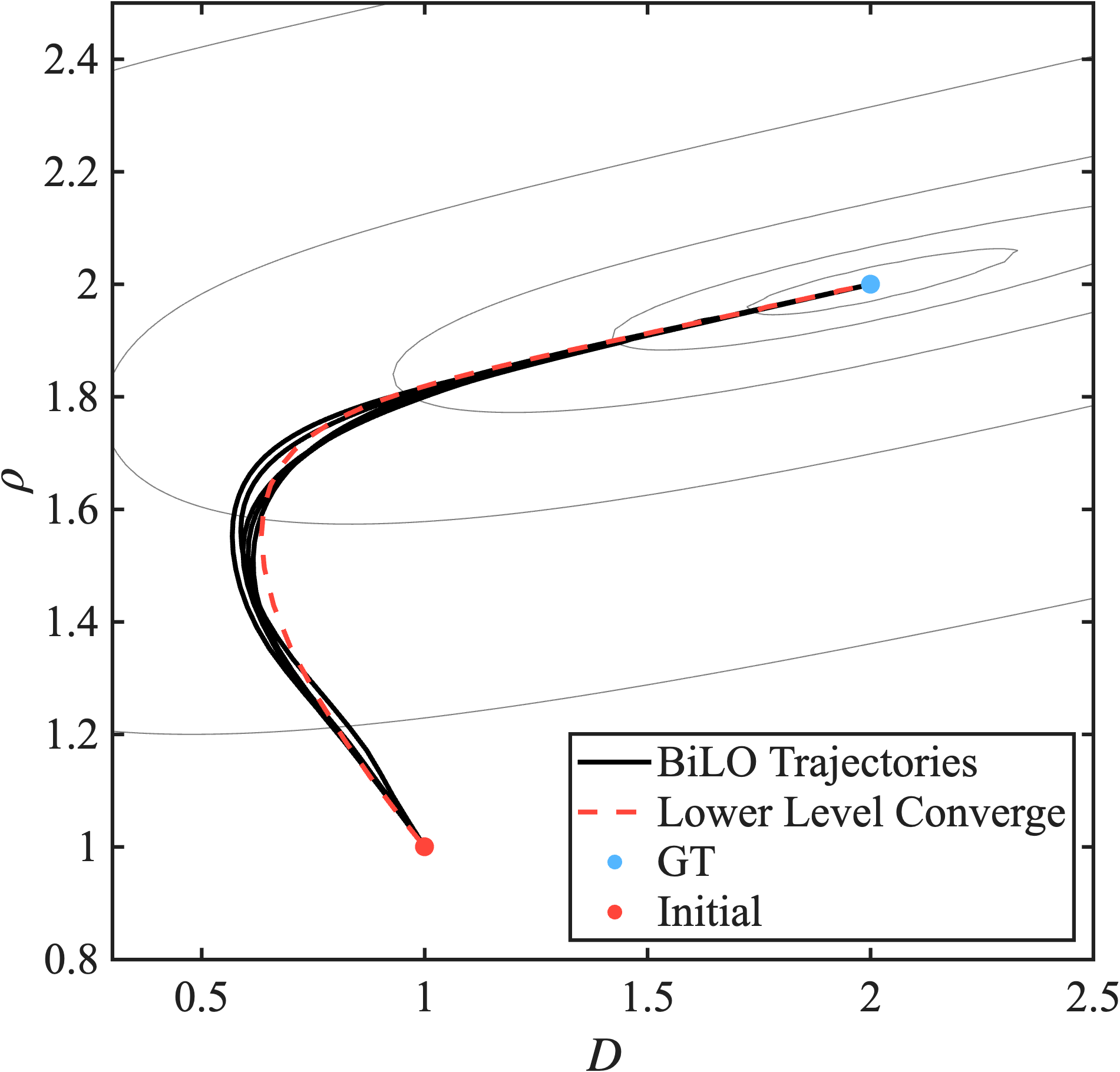}
  \caption{
Trajectories of the parameters \(D\) and \(\rho\) during fine-tuning.
The contours show the data-misfit loss in log scale computed using a finite-difference solution and serve as a reference loss landscape.
The dashed line denotes the trajectory obtained by solving the lower-level problem to a small tolerance before updating the PDE parameters, corresponding to a reference gradient descent path.
The black lines show BiLO trajectories from simultaneous gradient descent, which follow the reference path.
}
   \label{f:fktraj}
 \end{figure}

We show the trajectories of the parameters $D$ and $\rho$ during the fine-tuning process in Fig.~\ref{f:fktraj} without noise.
\change{
The contours are the data loss in log scale using the FDM solution for each parameter pair $(D,\rho)$. They serve as a reference loss landscape for visualization. 
The dashed red line shows the trajectory obtained by solving the lower level problem to a small tolerance ($10^{-4}$) before each update of the PDE parameters at the upper level.
This trajectory can be considered as the gradient descent path on the reference loss landscape.
Each BiLO trajectory (black line) corresponds to a different random initialization of the neural network and is obtained by the simultaneous gradient descent. 
Because the lower-level problem is not solved to convergence at each iteration, the resulting descent directions at each step contain small fluctuations; nevertheless, the BiLO trajectories closely follow the reference path.
}

\subsubsection{Estimation under Noisy Data}
We solve the inverse problem using different methods with different noise levels. 
Due to the presence of noise, the minimizer of the PDECO is no longer the ground truth parameters that generate the data.
We evaluate three approaches: (1) BiLO, (2) PINNs with different $\wdat$, and (3) Neural Operators with varying pretraining datasets. 
We show the mean and standard deviation of various metrics: the relative error of the inferred parameters $D$ and $\rho$ with respect to the GT, the relative L2 error of $\unn$ compared to $\ufdm$.

\begin{figure}[!h]
  \centering
  \includegraphics[keepaspectratio,width=\textwidth]{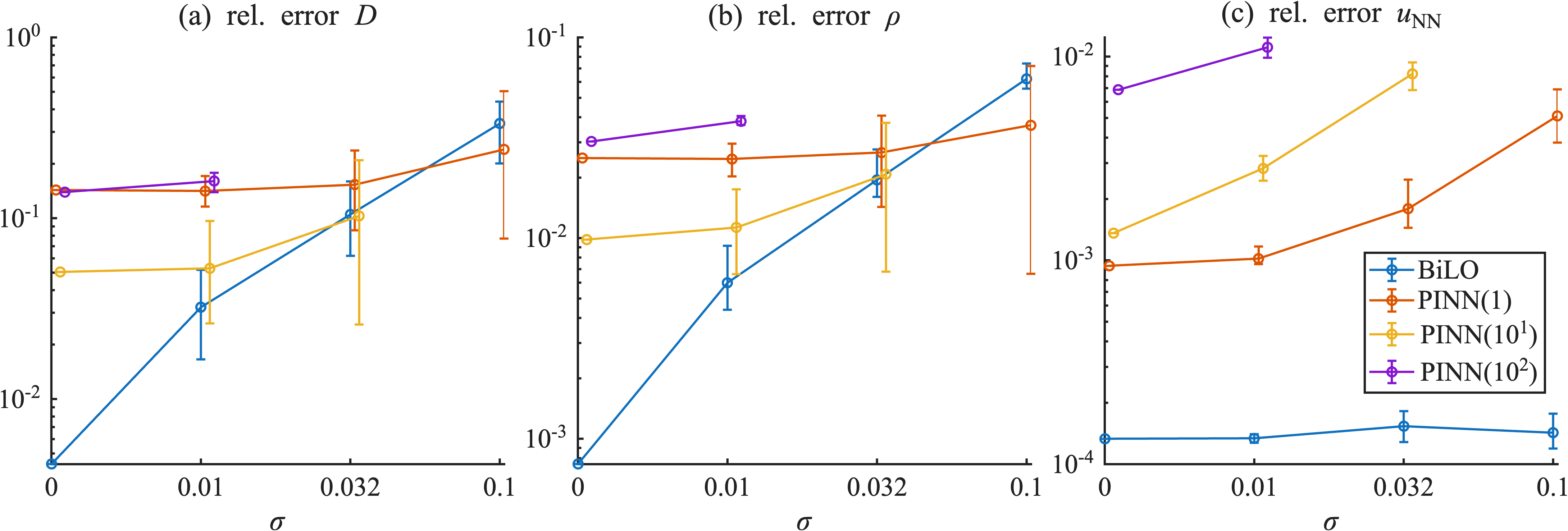}
  \caption{Comparison of performance (a,b) relative error of inferred parameters $D$ and $\rho$, (c) relative L2 error of $u_{NN}$ compared to $\ufdm$
  for $\sigma^2 = 0, 10^{-4}, 10^{-3}, 10^{-2}$
  across different methods: BiLO and PINN (with $\wdat = 1, 10, 100$).}
  \label{f:fknoize}
\end{figure}

\paragraph{Comparison with PINN}
Figure~\ref{f:fknoize} summarizes the performance of BiLO and PINN across varying noise levels and data loss weights. 
As expected, increasing the noise level generally leads to higher errors in the inferred parameters for all methods.
The results for PINN are highly sensitive to the choice of the data loss weight $\wdat$ and depend non-monotonically on $\wdat$ when the noise level is small. When the noise level is high, the accuracy deteriorates significantly and smaller $\wdat$ yield better, but  still limited accuracy. The weight $\wdat=10^3$ and noise $\sigma = 0.1$ result in unphysical solutions, such as negative values for $D$, and are therefore omitted from the plot.
Across all noise levels, BiLO consistently outperforms PINN in terms of parameter accuracy, with especially pronounced improvements for low or zero noise—achieving up to an order of magnitude lower error (note the logarithmic scale on the y-axis). 
In contrast, BiLO demonstrates robust performance in both parameter inference and solution accuracy, maintaining low error even as the noise level increases.
For PINN, the accuracy of the solution decreases as noise level increases. The deterioration becomes more pronounced as $\wdat$ increases. For the accuracy of the solution, BiLO is robust against noise levels.

\paragraph{Comparison with Neural Operators}
For the NO, we use the DeepONet architecture \cite{luLearningNonlinearOperators2021} as an example, which is shown to have comparable performance with FNO \cite{liFourierNeuralOperator2021,luComprehensiveFairComparison2022}.
In this experiment, we first train the NO using numerical PDE solutions corresponding to different values of $D$ and $\rho$, and then we use the NO as a surrogate and use gradient-based optimization to infer the parameters of the PDE.
We show that the quality of the inferred parameters depends on the training data used to train the NO.
 
We use the notation $a:h:b$ to denote an array from $a$ to $b$ with step $h$.
We consider the following 3 datasets for pretraining, where
the PDE parameters are sampled with different ranges and different resolutions:
\begin{itemize}[leftmargin=*, itemsep=0pt]
  \item Coarse: $D = 0.8:0.05:3$, $\rho = 0.8:0.05:3$.
  \item Dense: $D = 0.8:0.02:3$, $\rho = 0.8:0.02:3$.
  \item Out-of-distribution (OOD): $D = 0.8:0.02:3$, $\rho = 0.8:0.02:1.8$.
\end{itemize}
In the ``Coarse'' dataset, the parameters are sampled with a coarse grid;
In the ``Dense'' dataset, the parameters are sampled with a fine grid. 
In the ``OOD'' dataset, the parameters are sampled with a fine grid, but the ground truth $\rho$ is outside the range of the training data.

\begin{figure}[!h]
  \centering
  \includegraphics[keepaspectratio,width=\textwidth]{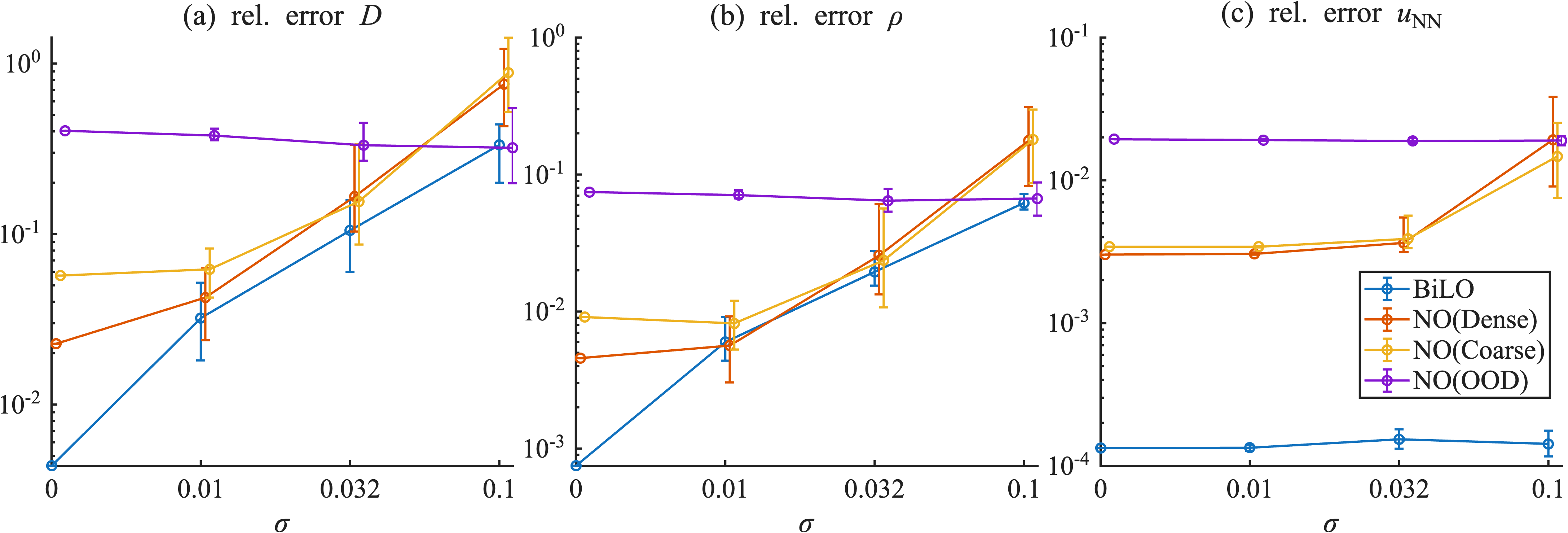}
  \caption{Comparison of performance (a,b) relative error of inferred parameters $D$ and $\rho$, (c) relative L2 error of $u_{NN}$ compared to $\ufdm$
  for $\sigma^2 = 0, 10^{-4}, 10^{-3}, 10^{-2}$
  across different methods: BiLO and NO (with different pretrain datasets).}
  \label{f:fknoize2}
\end{figure}

Figure~\ref{f:fknoize2} illustrates that overall, BiLO achieves more accurate parameter estimation, better solution accuracy compared to NO-based methods. 
The performance of NO is dependent on the choice of pretraining dataset. In particular, the NO trained on out-of-distribution data exhibits degraded performance, as the inferred parameters fall outside the support of the training distribution, resulting in relatively large errors in both the estimated parameters and the reconstructed solution. 
The accuracy of NO shows less sensitivity to noise levels than PINN, consistent with its role as a surrogate solver.

\subsection{Elliptic Equation with Singular Forcing}
\label{ss:pointproc}

We consider the following elliptic equation with singular forcing, which models the steady-state spatial distribution of mRNA molecules resulting from gene expression in a cell \cite{milesInferringStochasticRates2024}:
\begin{equation} \label{eq:pointproc}
  \begin{cases}
    \Delta u + \lambda \delta (x-z) - \mu u = 0 \quad\text{in } \Omega\\
    u = 0 \quad \text{ on } \partial\Omega
  \end{cases}
\end{equation}
Here, $u(x)$ is the intensity measure of a spatial Poisson point process describing the location of mRNA particles in a simplified 1D domain, \change{and $\delta(x - z)$ denotes the Dirac delta function representing a point source at the gene site $z$}.
The parameter $\lambda$ is the dimensionless transcription (birth) rate of mRNA, $\mu$ is the degradation rate, and the boundary conditions correspond to nuclear export. This formulation is motivated by inferring the dynamics of gene expression from static images of single cells.

Given $M$ snapshots of particle locations $q_i^j$ for $i=1,...,N_j$ and $j=1,...,M$,
we aim to infer the parameters $\lambda$ and $\mu$,
by minimizing the negative log-likelihood function, which we refer to as the ``data loss'':
\begin{equation}
  \min_{\lambda,\mu}  M\int_\Omega u(x) dx - \sum_{j=1}^M \sum_{i=1}^{N_j} \log u(q_i^j)
  \label{eq:nll}
\end{equation}

To solve this example in 1D with $\Omega = [0,1]$, the equation \eqref{eq:pointproc} can be written as a elliptic interface problem:
\begin{equation} \label{eq:interface}
  \begin{cases}
    \Delta u - \mu u = 0\\
    u(0) = u(1) = 0 \\
    u^+(z) = u^-(z)\\
    u_x^+(z) - u_x^-(z) = -\lambda
  \end{cases}
\end{equation}
where the superscript $+$ and $-$ denote the limiting values from the right and left side of the interface at $z$, respectively. 
The solution is continuous, but its derivative is discontinuous at $z$.

We handle the singular forcing using the cusp-capturing PINN \cite{tsengCuspcapturingPINNElliptic2023},
which has been proposed to learn functions of the form $\tilde{u}(x,\phi)$ such that $u(x) = \tilde{u}(x,|x-z|)$.
The continuity condition is automatically satisfied, and the jump condition translates into an additional constraint in $\cF$: 
$$\partial_\phi \tilde{u}(z,0) = -\lambda.$$ 
The cusp-capturing PINN is parameterized by $\tilde{u}(x, \phi; \wnn)$, and to enforce the jump condition, we need the ``jump loss'':
\begin{equation}
  \mathcal{L}_{\rm jump}^{PINN}(W) = (\partial_\phi \tilde{u}(z, 0; \wnn) + \lambda)^2
\end{equation}
The total loss for the PINN will include the residual loss, the jump loss, and the data loss.

\begin{figure}[!h]
  \centering
  \includegraphics[width=0.7\linewidth,keepaspectratio]{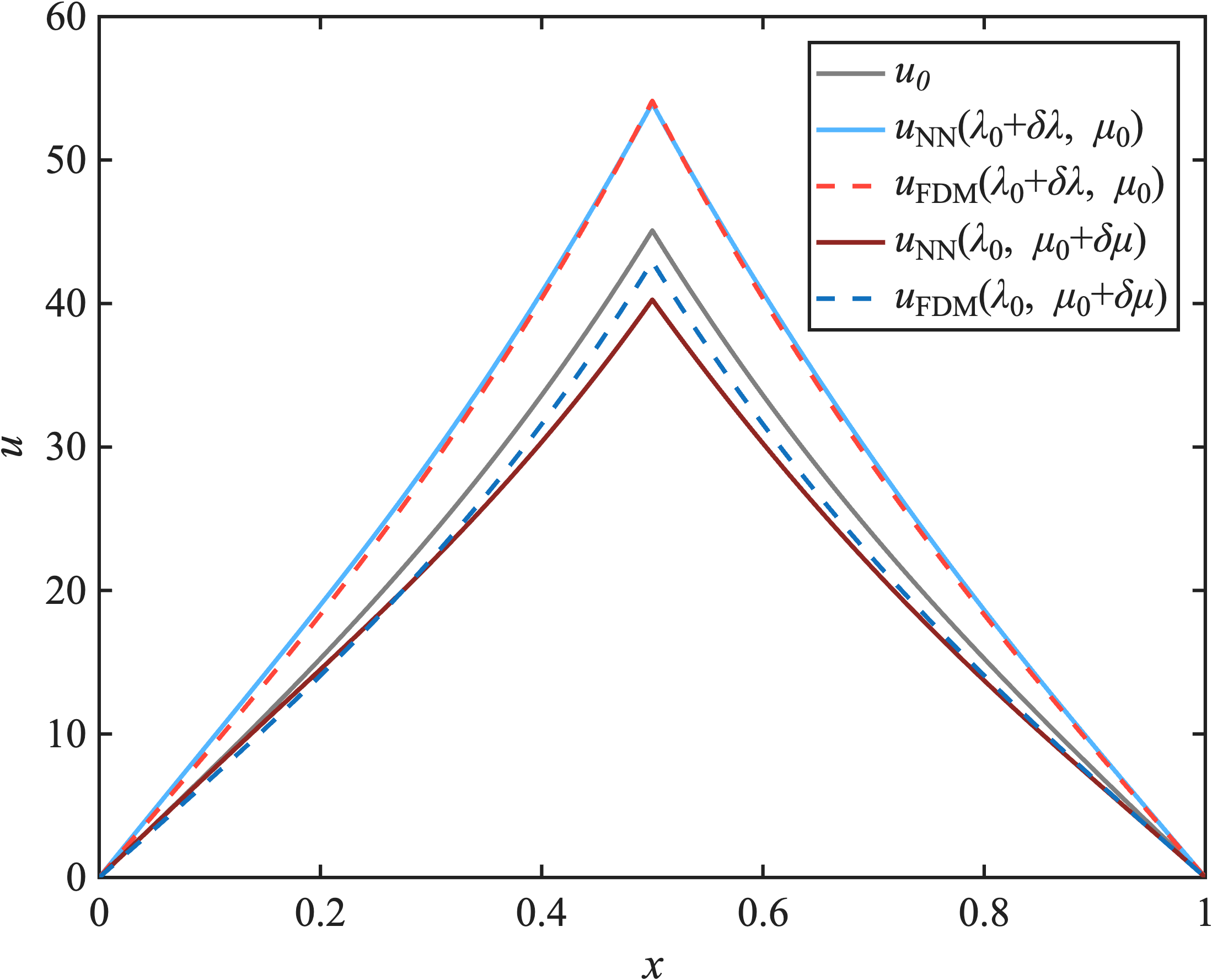}
  \caption{
Visualization of BiLO after pretraining at $\lambda_0 = 250$, $\mu_0 = 5$ and $z=0.5$, with perturbations $\delta\lambda = 100$ and $\delta\mu = 1$. Evaluating the pretrained operator \(\unn\) at nearby parameter values yields approximations to the corresponding FDM solutions \(\ufdm\).}
  \label{f:pointproc_var}
\end{figure}

In the BiLO framework, the local operator is parameterized as $\tilde{u}(x, \phi, \theta; \wnn)$, where $\theta = (\lambda, \mu)$. The ``jump loss'' is defined as
\begin{equation}
  \mathcal{L}_{\rm jump}(\theta,W) = (\partial_\phi \tilde{u}(z, 0, \theta; \wnn) + \lambda)^2
\end{equation}
and we also need the ``jump gradient loss'':
\begin{equation}
  \mathcal{L}_{\rm jgrad}(\theta, W) = \left(\nabla_\theta \partial_\phi \tilde{u}(z,0,\theta; \wnn)\right)^2
\end{equation}
\change{
The integral in the upper-level loss is approximated using the Simpson's rule with 100 uniform grid points.
The lower-level loss is given by $\mathcal{L}_{\rm res} + \mathcal{L}_{\rm jump} + \wdr (\mathcal{L}_{\rm rgrad} + \mathcal{L}_{\rm jgrad})$.
The upper-level loss is given by the data loss \eqref{eq:nll}
}
We visualize the local operator $u(x,\lambda, \mu;\wnn)$ in Fig.~\ref{f:pointproc_var}.
The objective function is more challenging than the MSE, as the values of $u$ are not available at the particle locations --  $u$ is only proportional to the histogram of the particle locations.
In this problem, the decay rate $\mu$ is typically on the order of 10, while the birth rate $\lambda$ is on the order of several hundreds, consistent with biologically plausible dynamics. To improve numerical conditioning during training, we reparameterize $\lambda = 100 \bar{\lambda}$ and learn the rescaled parameter $\bar{\lambda}$. In addition, we apply a final transformation $\tau(m,x) = m^2x(1 - x)$, where $m$ is the raw output of the neural network. This transformation enforces the boundary conditions and ensures non-negativity of the solution, which is necessary for evaluating $\log u$ in the likelihood.
The initial guesses are $\lambda_0 = 250$ and $\mu_0 = 5$, with a relative error of $50\%$ compared to the ground truth.
The particle positions are sampled using ground truth parameters $\lambda_{\rm GT} = 500$ and $\mu_{\rm GT} = 2.5$ with $M=100$. 

For experiments using neural operator, we consider two pre-training datasets: 
\begin{itemize}[leftmargin=*, itemsep=0pt]
  \item In-distribution (ID): $\lambda = 100:20:800$, $\mu = 2:1:20$.
  \item Out-of-distribution (OOD): $\lambda = 100:20:800$, $\mu = 3.5:0.5:10$.
\end{itemize}
The $\mu_{\rm GT}$ are outside the range of the training data in the OOD dataset.

\begin{figure}[!h]
  \centering
  \includegraphics[width=\linewidth,keepaspectratio]{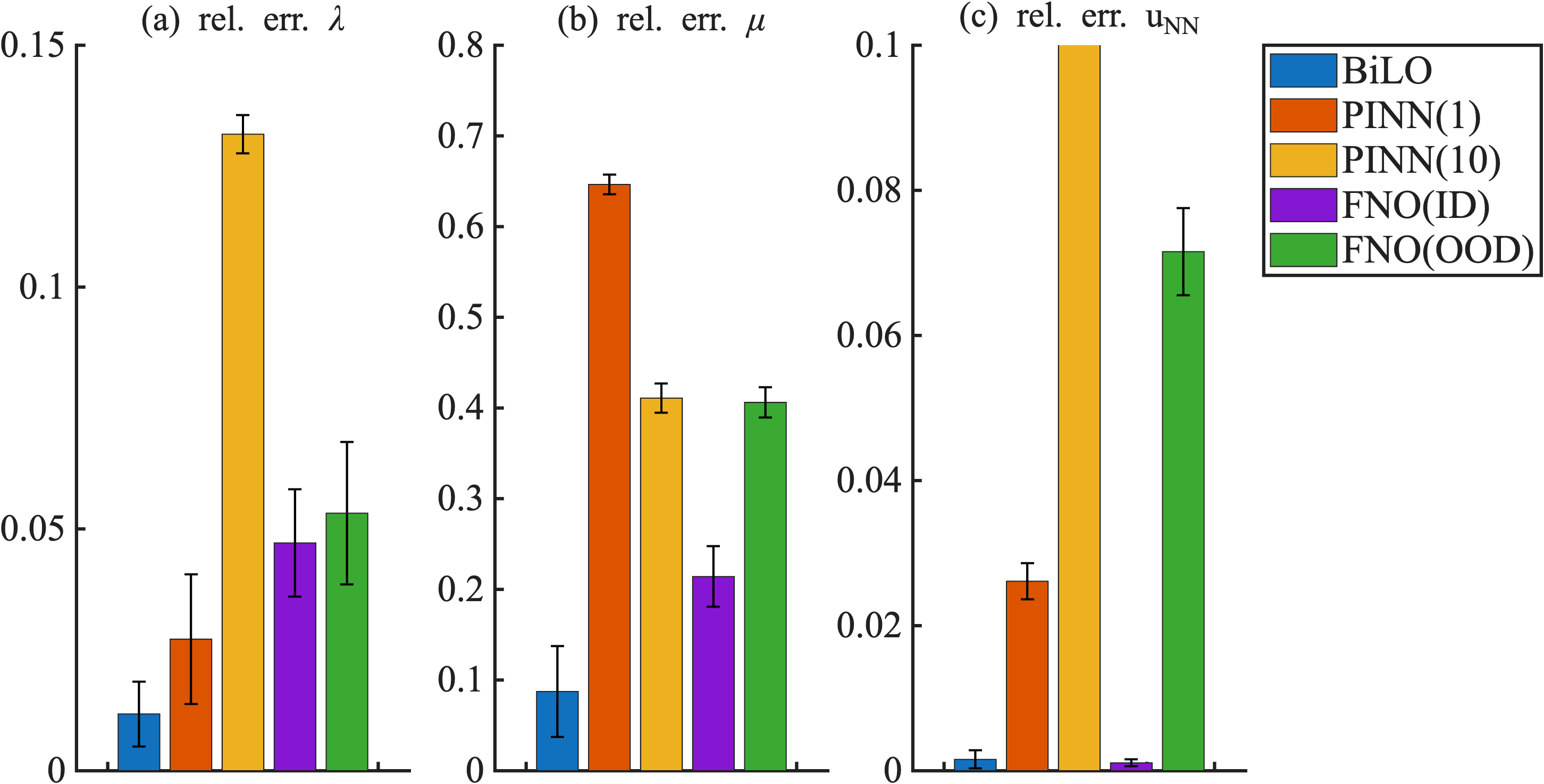}
  \caption{Comparison of performance metrics for BiLO and PINN ($\wdat$ = $1$ and $10$) and FNO (with different pretraining datasets)
  for the elliptic equation with singular forcing.}
  \label{f:pointproc}
\end{figure}

The results are shown in Fig.~\ref{f:pointproc}, and BiLO achieves significantly better performance than PINN ($\wdat$ = $1$ and $10$) and FNO (with different pretraining datasets).
For PINN, the results depend on the choice of $\wdat$. 
$\wdat = 10$ leads to larger error in $\lambda$ and overfitting, as indicated by the elevated relative error of $u_{NN}$ compared to $\ufdm$. However, $\wdat = 1$ leads to larger relative error in $\mu$.
For FNO, when pretrained on the ID dataset, although the relative errors in parameter estimation are larger than that of BiLO, the PDE solution accuracy remains comparable, with no signs of overfitting or underfitting. However, the OOD dataset results in larger parameter errors and a less accurate solution.

\subsection{Poisson Equation with Variable Diffusion}
\label{ss:varpoi}
We consider the following 1D Poisson equation
\begin{equation}
  \begin{cases}
    -(D(x) u'(x))' = f(x) \quad\text{in } [0,1]\\
    u(0) = u(1) = 0
  \end{cases}
  \label{eq:var}
\end{equation}
where $f(x) = \pi^2\sin(\pi x)$. 
We aim to infer the variable diffusion coefficient $D(x)$ such that $D(0) = D(1) = 1$.
The ground truth $D(x)$ is a ``hat'' function:
\begin{equation}
  D(x) = \begin{cases}
    1 + 0.5x, & \text{if } x \in [0,0.5)\\
    1.5 - 0.5x, & \text{if } x \in [0.5,1]
  \end{cases}
\end{equation}
\change{
The relative error of the initial guess in $L^\infty$ norm is around $33\%$.
Moreover, the initial guess is overly smooth and misses the kink at $x=0.5$, offering little information about the true profile.}

\begin{figure}[!h]
  \centering
  \includegraphics[keepaspectratio,width=0.7\linewidth]{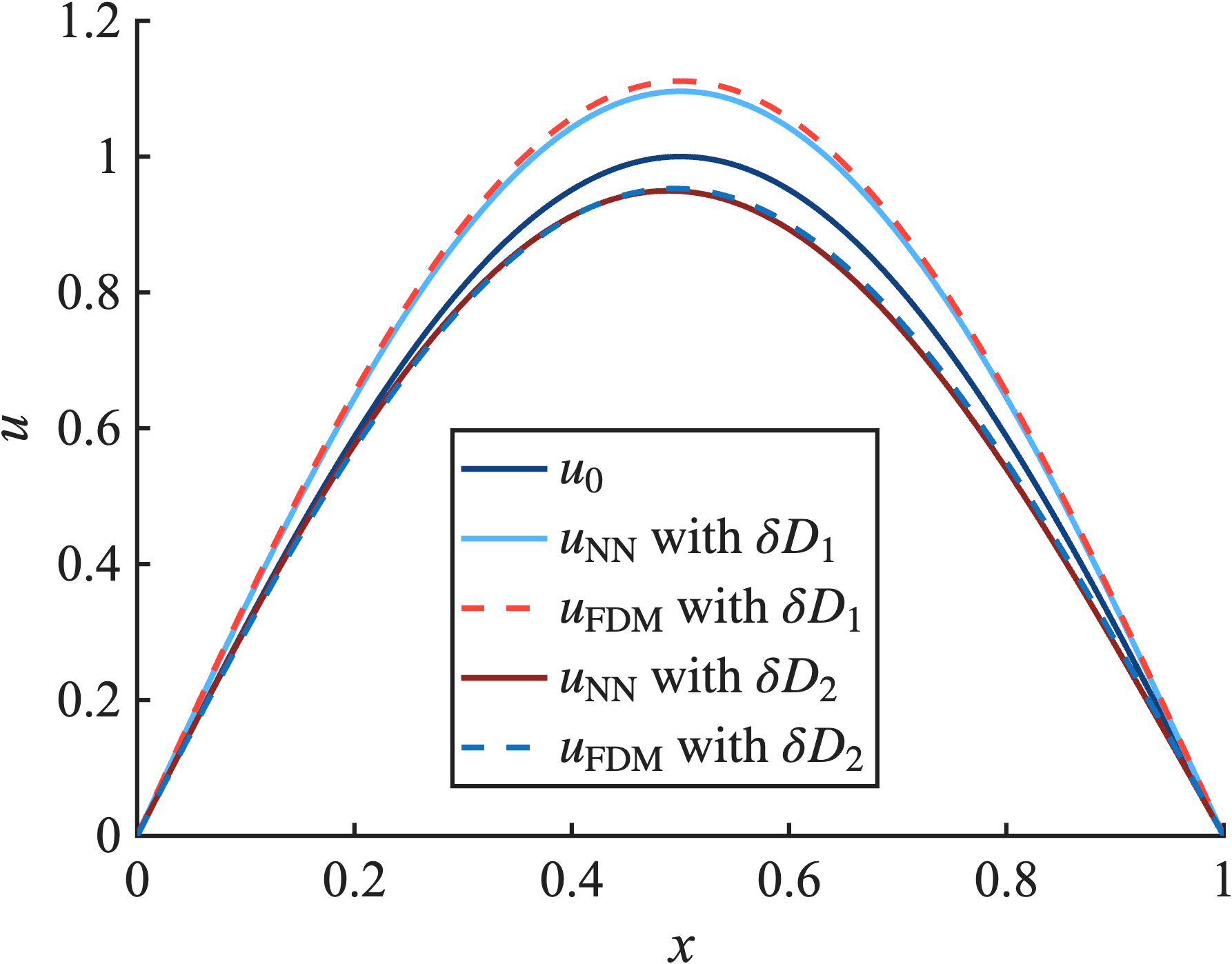}
  \caption{Visualizing the operator $u(x,D_0(x) + \delta D_i(x);\wnn)$,
  where $\delta D_1(x) = -0.1$ and $\delta D_2(x) = 0.1x$, after pre-training with $D_0(x) = 1$.
  The FDM solutions corresponding to $D_0(x) + \delta D_i(x)$ are also shown.
  }
  \label{f:varpoi_var}
\end{figure}

\paragraph*{Visualizing BiLO}
In Fig.~\ref{f:varpoi_var}, we visualize the local operator $u(x,z;\wnn)$ after pre-training with $D_0(x) = 1$. 
We consider the variation 
$\delta D_1(x) = -0.1$, and 
$\delta D_2(x) = 0.1x$
and evaluate the neural network at $u(x, D_0(x)+\delta D_i(x);\wnn)$ for $i=1,2$.
The FDM solutions of the PDE corresponding to $D_0(x)+\delta D_i(x)$ are also plotted.
We can see that the neural network provides a good approximation to the solution corresponding to $D_0(x)+\delta D_i(x)$.

\begin{figure}[!h]
  \centering
  \includegraphics[keepaspectratio,width=\linewidth]{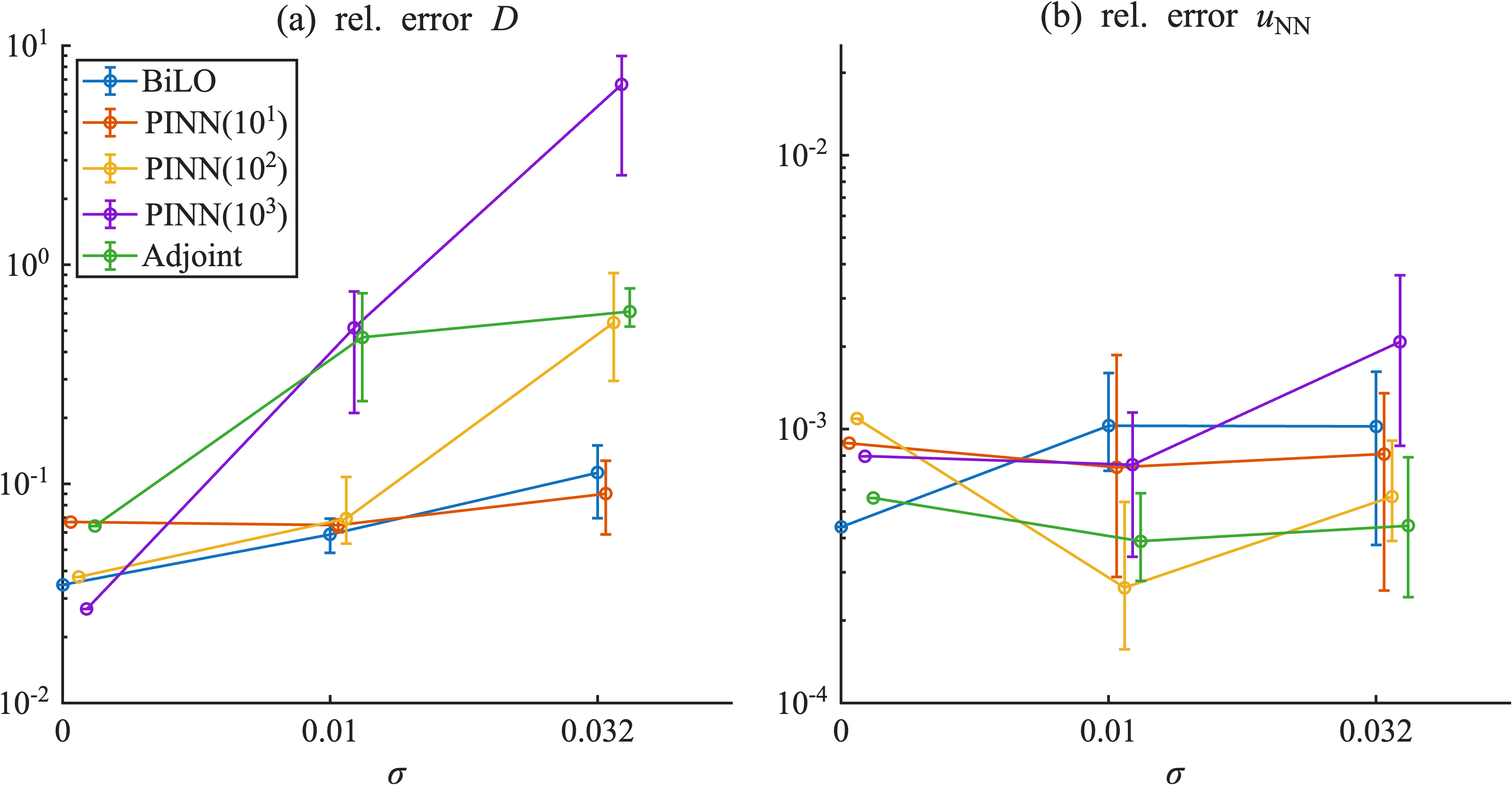}
  \caption{Comparison of performance metrics -- (a) relative error in $D$ and (b) relative error in $u$ -- for BiLO, PINN (with various $\wdat$), and the adjoint method}
  \label{f:varpoi_pinn}
\end{figure}

\paragraph*{Results}
We next estimate $D(x)$ in the presence of noise at levels $\sigma=0$, $0.01$, $0.03$ and use $\wreg = 10^{-3}$. 
%
We use BiLO, PINNs with $\wdat = 1, 10, 100$, and the adjoint method.
Figure~\ref{f:varpoi_pinn} shows that BiLO consistently produces more accurate estimates of the diffusion coefficient $D(x)$ across all noise levels. In contrast, the performance of PINN is highly sensitive to the choice of $\wdat$: larger values such as $\wdat = 10^3$ work well under noise-free conditions but lead to poor performance when the noise level increases; smaller values such as $\wdat = 10$ are more robust to noise but underfit when no noise is present. The reconstruction accuracy of the solution $u_{NN}$ is comparable between BiLO and PINN.
The adjoint method, which solves the PDE with high accuracy and is considered "exact" (hence no corresponding bar in (b)), reconstructs the diffusion coefficient less accurately than BiLO.

We also infer the variable diffusion coefficient $D(x)$ in the Poisson equation using a DeepONet.
The pretraining dataset is generated by solving the Poisson equation with 1000 samples of variable diffusion coefficients $D(x)$.
$D(x)$ is sampled from a Gaussian Random field on $[0,1]$, conditioned on $D(0)=D(1)=1$.
The covariance function is the Gaussian kernel, with variance 0.05 and different length scales $l = 0.2,0.3,0.4$. 
As $l$ increases, the samples of $D(x)$ become smoother.

\begin{figure}[!h]
  \centering
  \includegraphics[keepaspectratio,width=\linewidth]{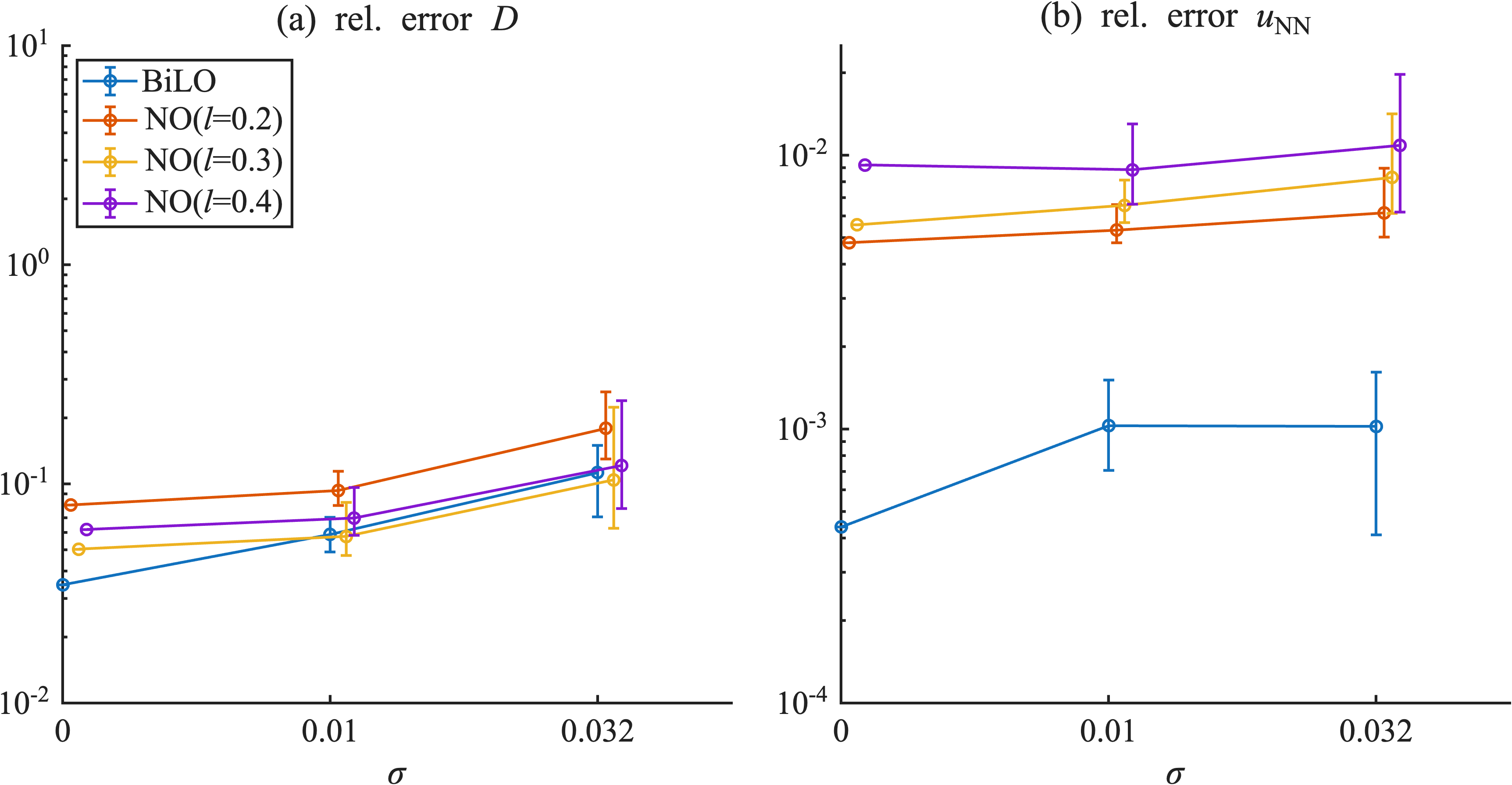}
  \caption{Comparison of performance metrics for different methods: BiLO and NO (with different pretrain datasets).}
  \label{f:varpoi_no}
\end{figure}

Figure~\ref{f:varpoi_no} shows that the performance of NO depends on the choice of pretraining dataset. Because of the ill-posedness, the inference of $D(x)$ is similar across the different methods, although BiLO is more accurate at smaller noise levels and at least as accurate as NO at larger noise levels. While the accuracy of NO in approximating the solution  remains relatively stable across different noise levels, it is less accurate than both PINN and BiLO. Among all methods, BiLO consistently achieves comparable or better inferences of the diffusion coefficient $D(x)$.

\begin{figure}[!h]
  \centering
  \includegraphics[keepaspectratio,width=\linewidth]{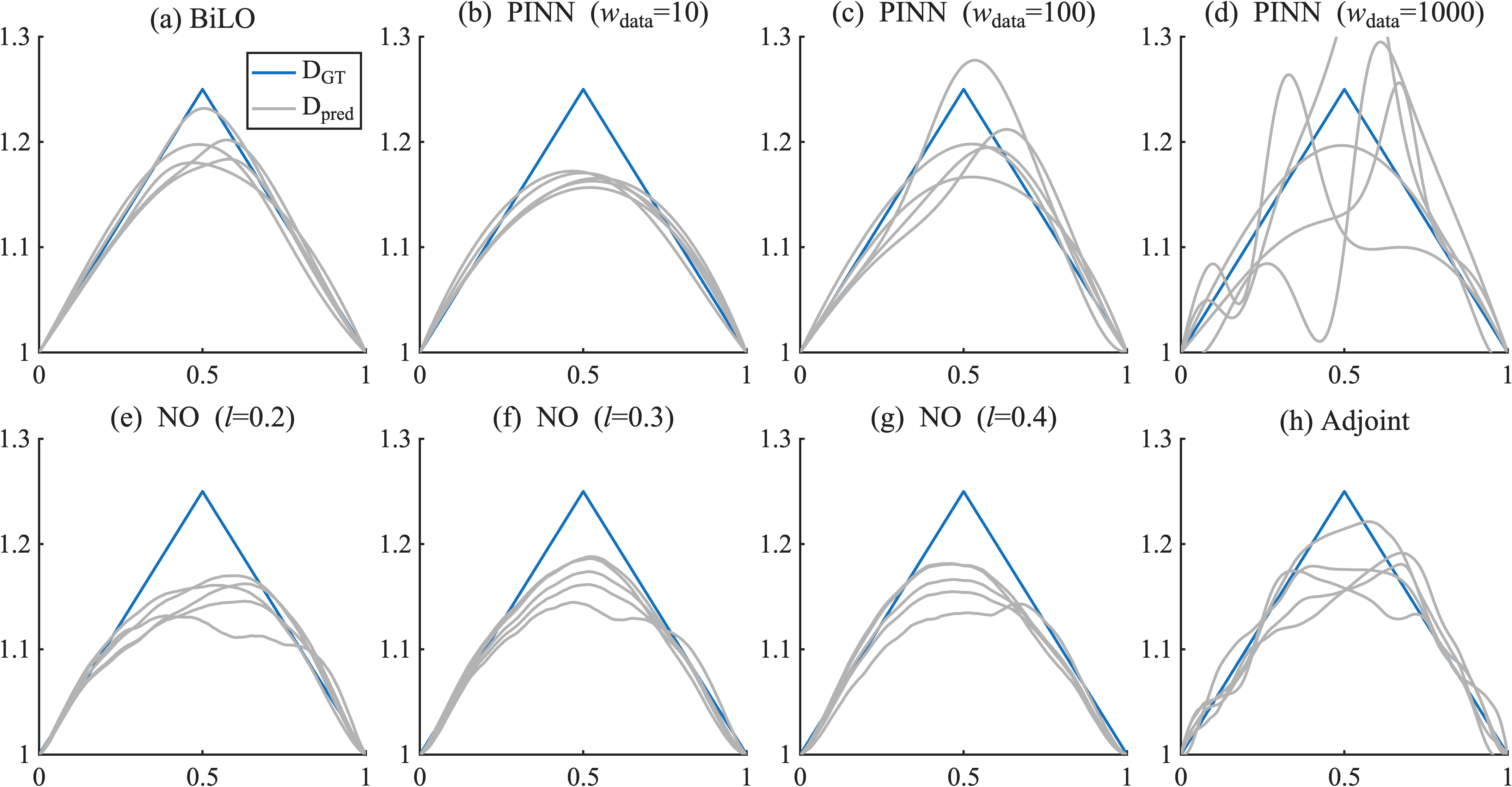}
  \caption{Examples of the inferred diffusion coefficient $D(x)$ from various methods with $\sigma=0.01$.}
  \label{f:varpoi}
\end{figure}

Figure~\ref{f:varpoi} illustrates qualitative differences in the inferred diffusion coefficient $D(x)$ across the methods. BiLO best captures the kink in the ground truth, leading to a more accurate reconstruction. In contrast, PINN with low $\wdat$ produces overly smooth estimates, while large $\wdat$ leads to oscillatory artifacts due to overfitting. The NO reconstructions, regardless of the length scale used in pretraining, tend to be smooth and lack sharp features. The adjoint method yields slightly more oscillatory reconstructions compared to BiLO.

\subsection{Inferring the Initial Condition of a Heat Equation}
\label{ss:heat}
\begin{figure}[!ht]
  \centering
  \includegraphics[keepaspectratio,width=0.7\linewidth]{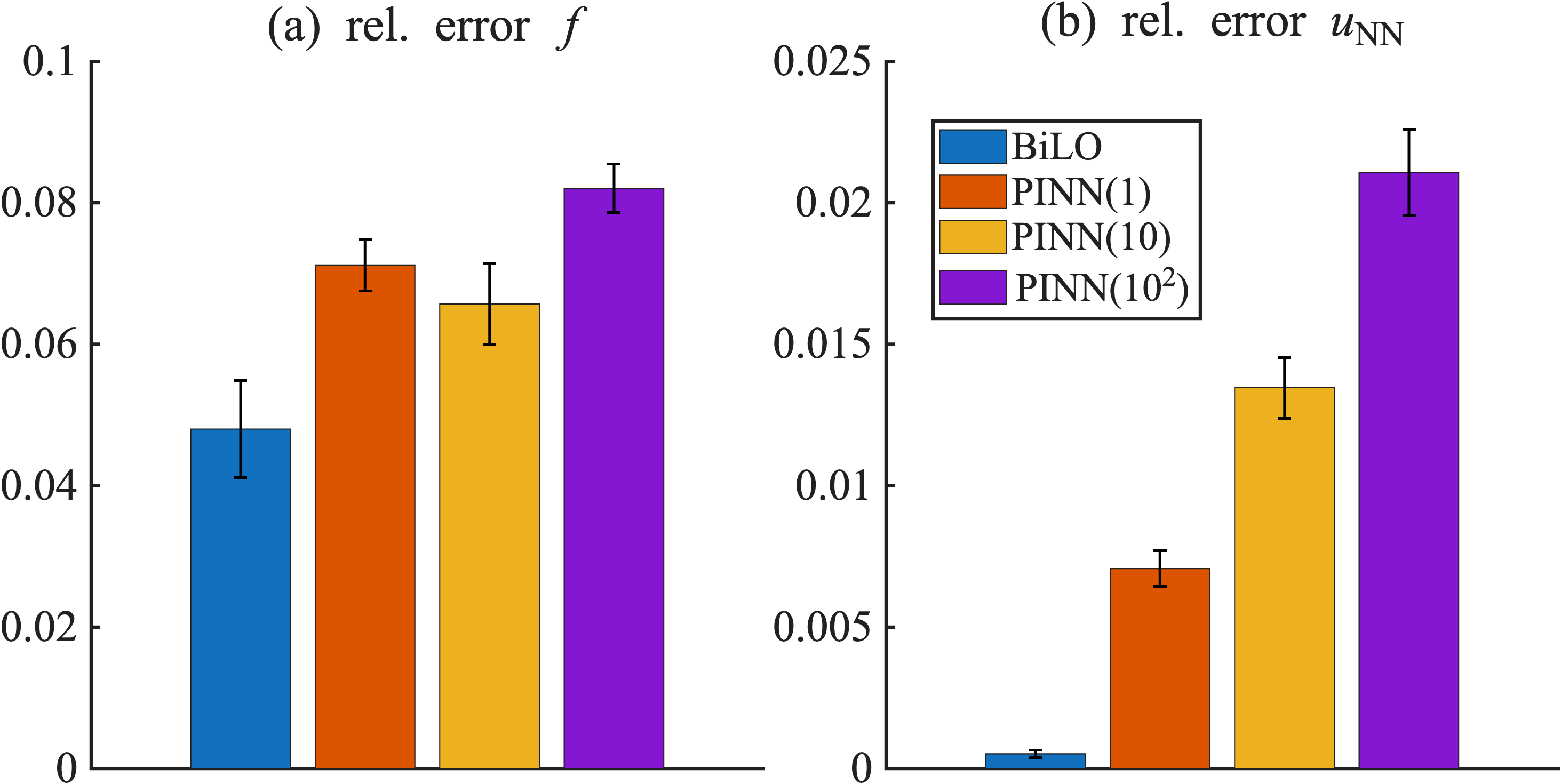}
  \caption{Comparison of the BiLO and PINN (with various $\wdat$) for inference of the initial condition of the heat equation}
  \label{f:heat}
\end{figure}

In this example, we aim to infer the initial condition of a 1D heat equation from the final state.
Consider the heat equation
\begin{equation}
  \begin{cases}
    u_t(x,t) = D u_{xx}(x,t)\\
    u(x,0) = f(x)\\
    u(0,t) = u(1,t) = 0  \\
  \end{cases}
\end{equation}
on $x\in[0,1]$ and $t \in[0,1]$, with fixed diffusion coefficient $D=0.01$, and unknown initial condition $f(x)$, where $f(0) = f(1) = 0$.
Our goal is to infer the initial condition $f(x)$ from observation of the final state $u(x,1)$.
We set the ground truth initial condition $f_{\rm GT}$ to be the hat function
\begin{equation}
  f_{\rm GT}(x) =
  \begin{cases}
    2x, & \text{if } x \in [0,0.5)\\
    2 - 2x, & \text{if } x \in [0.5,1]
  \end{cases}
\end{equation}
We set the initial guess $f_0(x) = \sin(\pi x)$.
\change{The initial guess has a relative $L^\infty$ error of about 50\%.
In addition, its smooth sinusoidal shape does not capture the nondifferentiability at x=0.5.}
To evaluate the performance of the estimated initial condition $f$,
we use the $L^2$ norm of the estimated initial condition and the ground truth initial condition,
which are evaluated at 1001 evenly spaced points in the spatial domain.
We consider the case with noise $\epsilon \sim N(0,0.001)$.
Due to the ill-posedness of the inverse problem, we need to regularize the problem by penalizing the 2-norm of the derivative of the unknown function with $\wreg = 10^{-2}$. The results, shown in Fig. \ref{f:heat}, show that BiLO outperforms PINNs with various values of $\wdat$.


\subsection{Inviscid Burgers' Equation}\label{ss:burger}

We consider an inverse problem governed by an inviscid Burgers' equation on the domain $x \in [0,1]$ and $t \in [0,1]$.

\begin{equation}
  \begin{cases}
    u_t + a uu_x = 0, \\
    u(x,0) = f(x), \\
    u(0,t) = u(1,t) = 0
  \end{cases}
\end{equation}
where $a = 0.2$. 
We infer the initial condition $f$ from the observational data at $t=1$.
The numerical solutions are computed by using the Godunov scheme. 
The inviscid Burgers' equation is a hyperbolic PDE, and the solution can develop shocks and rarefaction waves.

\begin{figure}[!h]
  \centering
  \includegraphics[keepaspectratio,width=\linewidth]{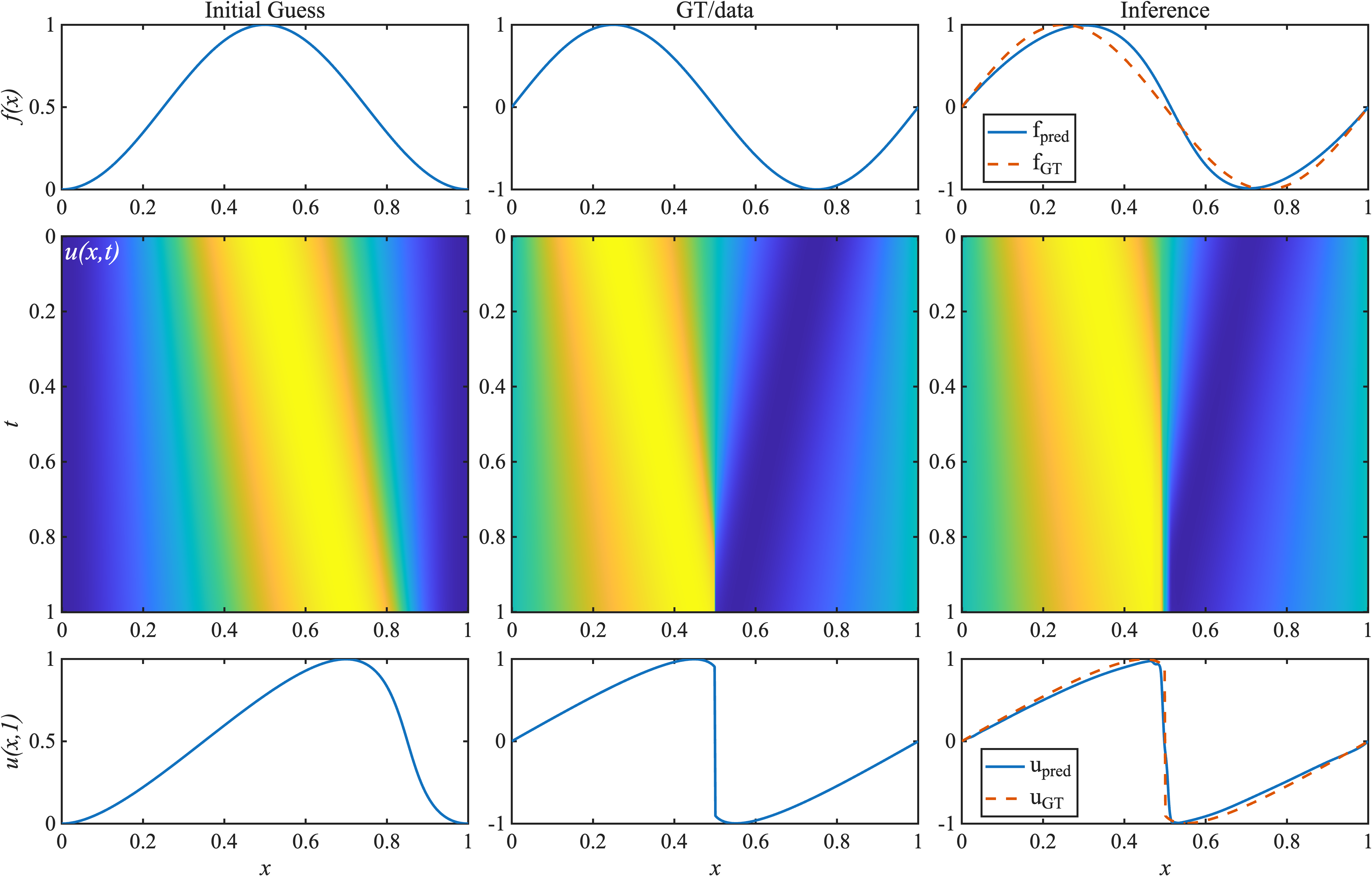}
  \caption{Example of inferring the initial condition of the inviscid Burgers' equation using data at $t=1$. 
  The initial guess is used to pre-train the network. The solution at $t=1$ of the GT is the data for inference. First column: initial guess, second column: ground truth, third column: inferred initial condition. First row: initial condition, second row: solution $u(x,t)$, third row: solution $u(x,1)$.}
  \label{f:burger1}
\end{figure}

In Fig.~\ref{f:burger1}, we show the initial guess in the first column, the ground truth in the second column, and the inference results by BiLO ($\wreg=10^{-3}$) in the third column.
The first row shows the initial condition $f(x)$, the second row shows the solution $u(x,t)$ on the domain $x \in [0,1]$ and $t \in [0,1]$, and the third row shows the solution $u(x,1)$.
For inference, only the solution at $t=1$ of the ground truth is provided.
\change{The solution of the initial guess remains smooth over the entire time interval $t\in[0,1]$, with no discontinuity forming.
In contrast, the ground-truth initial condition shown in the second column evolves into a solution with a discontinuity over time.
The inferred initial condition and solution shown in the third column recover this non-smooth behavior, demonstrating that BiLO can capture discontinuity despite starting from a smooth initial guess.}
Additional examples are shown in \ref{ap:ss:burger}.

\subsection{Darcy Flow in 2D}\label{ss:2dpoisson}

The setup of this experiment is similar to the steady state Darcy flow inverse problem in \cite{liPhysicsInformedNeuralOperator2024}. We consider the following 2D Poisson equation with variable diffusion coefficient $A(\bx)$ in the unit square domain $\Omega = [0,1]\times[0,1]$ with Dirichlet boundary condition:
\begin{equation}
  \begin{cases}
    - \nabla \cdot (A(\vb{x}) \nabla u(\vb{x})) = f(\vb{x}) & \text{in } \Omega\\
    u(\vb{x}) = 0, \quad \text{on } \partial \Omega
  \end{cases}
\end{equation}
Our goal is to infer the variable diffusion coefficient $A(\vb{x})$ from the solution $u(\vb{x})$.
Let $\phi(\vb{x})$ be samples of a Gaussian random field (GRF) with mean 0 and squared exponential (Gaussian) covariance structure 
$$C(\vb{x},\vb{y}) = \sigma \exp(-||\vb{x}-\vb{y}||^2/\lambda^2),$$ 
where the marginal standard deviation $\sigma=\sqrt{10}$ and the correlation length $\lambda=0.01$ \cite{constantineRandomFieldSimulation2024}.
We generate the initial guess $A_0(\vb{x}) = s(\phi_0(\vb{x})) \times 9 + 3$, where $\phi_0(\vb{x})$ is a sample of the GRF, and $s(x) = 1/(1 + e^{-x})$ is the logistic function.
We consider the ground truth diffusion coefficient to be a piecewise constant function with high and low diffusivity:
$A_{\rm GT}(\vb{x}) = 12$ if $\phi_{\rm GT}(\vb{x}) > 0$ and $A_{\rm GT}(\vb{x}) = 3$ otherwise, where $\phi_{\rm GT}$ is another sample of the GRF. 
\change{The neural network representation of $A(\vb{x})$ is smooth by construction; hence, we approximate the discontinuous ground truth diffusion coefficient by thresholding the inferred field at 7.5 \cite{liPhysicsInformedNeuralOperator2024}.}
The corresponding solution of $A_{\rm 0}$ and $A_{\rm GT}$ are denoted as $u_0$ and $u_{\rm GT}$.
Following \cite{liPhysicsInformedNeuralOperator2024}, we use the total variation regularization $|\nabla A|$ with weight $\wreg=10^{-9}$.
\begin{figure}[!h]
  \centering
  \includegraphics[keepaspectratio,width=\linewidth]{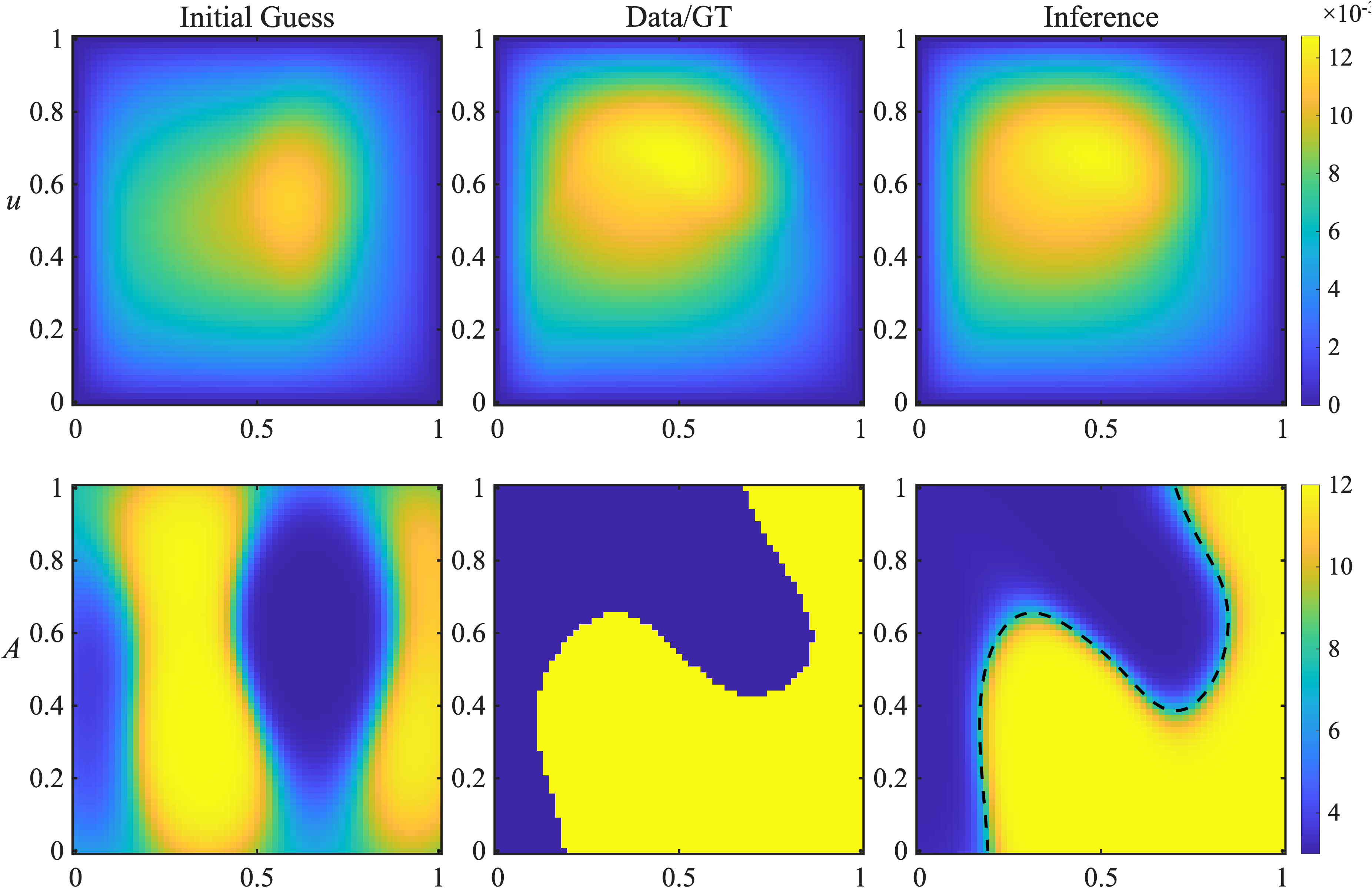}
  \caption{Example of inferring the variable diffusion coefficient in the 2D Darcy flow problem. The network prediction achieves a relative MSE of 1.7\% for $u$. Thresholding the inferred \(A(\vb{x})\) at 7.5 (dashed line) recovers the discontinuous ground-truth structure with a classification accuracy of 96.1\%}
  \label{f:darcy1}
\end{figure}

Figure~\ref{f:darcy1} shows an example of the results of the inverse problem. 
The first and the second row show $u$ and $A$, respectively.
The first column shows the initial guess, the second column shows the ground truth, and the third column shows the inference results by BiLO.
\change{The dashed line marks the 7.5 level used to classify regions of high and low diffusivity, yielding a classification accuracy of 96.1\%, which is comparable to the results (97.1\% classification accuracy)} from the Physics-informed Neural Operator (PINO) in \cite{liPhysicsInformedNeuralOperator2024}, which requires pretraining a FNO with a synthetic dataset, and instance-wise fine-tuning with physics-informed loss. In our method, we only need to pretrain the BiLO with a single initial guess, which is visually different from the ground truth. Additional examples are shown in \ref{ap:ss:2dpoisson}.

\subsection{Glioblastoma (GBM) Inverse Problem}
In this section, we consider a real-world application of BiLO for a patient-specific parameter estimation of GBM growth models using patient MRI data in 2D.
The challenge lies in the high noise levels and the potential model mis-specification, as the Fisher-KPP PDE likely does not fully capture the complexities present in the tumor MRI data.
The setup of the problem follows \cite{zhangPersonalizedPredictionsGlioblastoma2025,balcerakIndividualizingGliomaRadiotherapy2024,ezhovLearnMorphInferNewWay2023,scheufeleFullyAutomaticCalibration2021}. 
\paragraph{Tumor Growth and Imaging Model}
Let $\Omega$ be the brain region in 2D based on MRI images.
The normalized tumor cell density is $u(\vb x, t)$.
\begin{equation} \label{eq:pde1}
  \begin{cases}
    \pdv{u}{t} = D \bar{D} \grad \cdot (P(\vb x) \grad u) + \rho \bar{\rho} u (1-u) \quad \text{in }  \Omega \\
    \grad u \cdot \vb n = 0 \quad \text{on } \partial \Omega 
  \end{cases}
\end{equation}
where $P$ depends on the tissue distribution (e.g., white and grey matter) and is obtained from the MRI data, $\bar{D}$ and $\bar{\rho}$ are known patient-specific characteristic parameters based on the data, and $D$ and $\rho$ are the unknown nondimensionalized parameters that we aim to infer from the data.

We consider two regions of interest in the tumor, the whole tumor (WT) region and the tumor core (TC) region.
Let $\hat{\by}^{s}$, $s\in \{\rm WT, TC\}$ be indicator function of the WT and TC regions, which can be obtained from the MRI data and serves as the observational data in the inverse problem.
We assume that the segmentations are the regions in which the tumor cell density $u$ at nondimensional $t=1$ lies above a certain threshold $u_c^s$:
$$\by^s(\bx) = s(20(u(\bx,1) - u_c^s)),$$
where $s$ is the logistic function. 
The predicted segmentation depends on the solution of the PDE, and thus on the parameters $D$, $\rho$ and $u_c^s$.
In the end, we aim to minimize the relative error between the predicted and the observed segmentations:
\begin{equation}
  \min_{D, \rho, \ucwt, \uctc} {\sum_{s\in \{WT,TC\}}}||\by^{s} - \hat{\by}^{s}||_2^2/||\hat{\by}^{s}||_2^2 
\end{equation}

\paragraph*{Results}
In this inference problem, the ground truth values for the parameters $D$, $\rho$, $\ucwt$, and $\uctc$ are not available. Therefore, we evaluate the quality of the inferred parameters by comparing the predicted segmentations $\hat{\by}$ with the observed segmentation $\by$ data using the DICE score, which is defined as $2 \langle\by,\hat{\by}\rangle / (\|\by\|_1 + \|\hat{\by}\|_1)$ \cite{diceMeasuresAmountEcologic1945}.
DICE is a standard metric in medical image segmentation that quantifies the overlap between two binary masks \cite{lipkovaPersonalizedRadiotherapyDesign2019,zhangPersonalizedPredictionsGlioblastoma2025}.

The predicted segmentations are obtained by thresholding the tumor cell density at the inferred thresholds $u_c^{WT}$ and $u_c^{TC}$. These densities can be computed either from the numerical PDE solution $\ufdm$ or the neural network surrogate $\unn$, resulting in predicted masks denoted by $\by^s_{\rm FDM}$ and $\by^s_{\rm NN}$, respectively. We define $\mathrm{DICE}_m$, $m \in \{\text{NN}, \text{FDM}\}$, as the average DICE score across the WT and TC regions using the corresponding predicted segmentation.

\begin{figure}[!h]
  \centering
   \includegraphics[keepaspectratio,width=\linewidth]{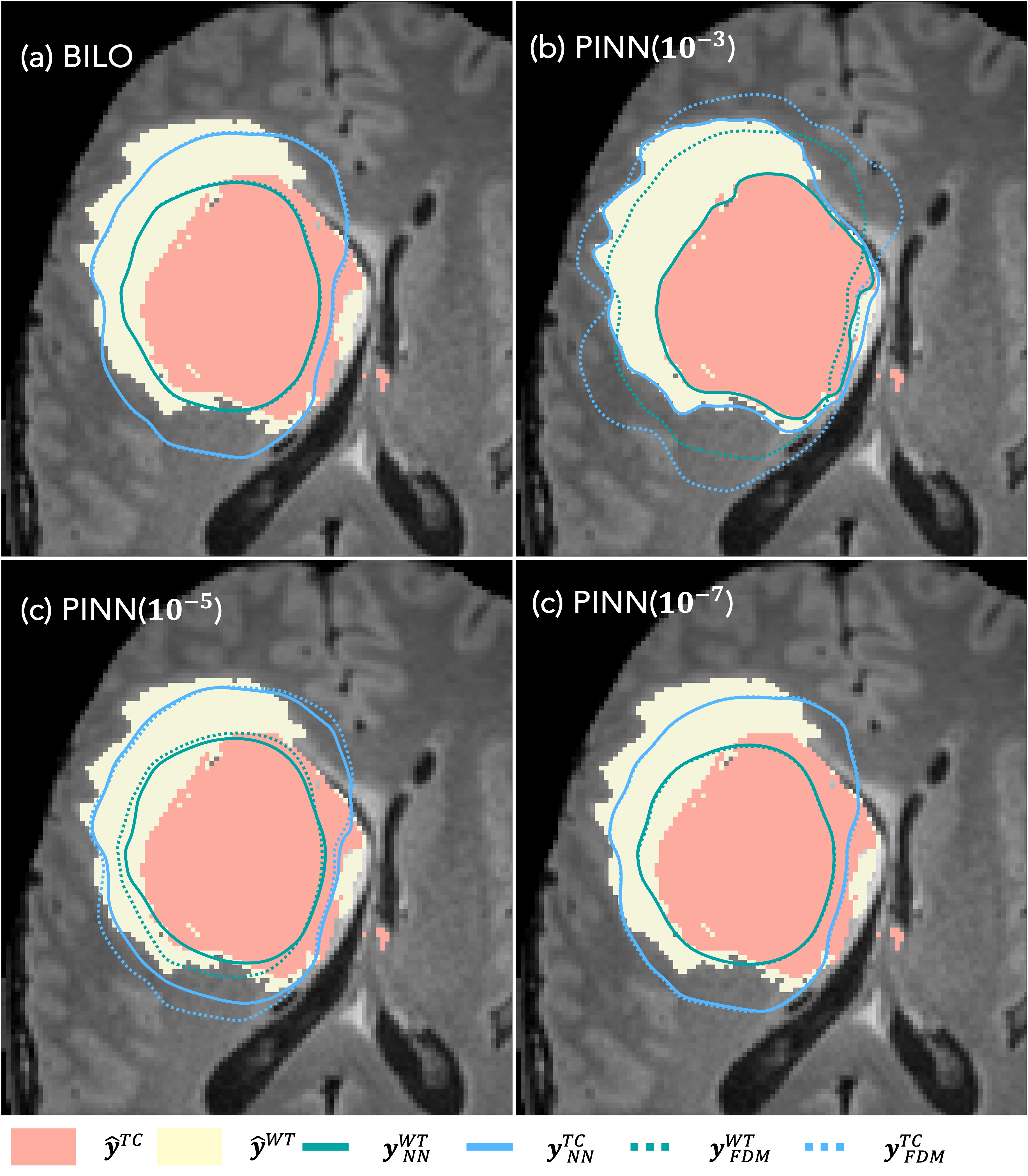}
  \caption{ Predicted segmentation using PINN with $\wdat$ = $10^{-3}$,  $10^{-5}$,  $10^{-7}$ and  BiLO.
  The solid and dashed contours are the predicted segmentation based on $\unn$ and $\ufdm$. BiLO gives almost overlapping contours, suggesting high accuracy of $\unn$.}
  \label{f:gbm}
\end{figure}

Table~\ref{t:gbm} reports the relative errors of $\unn$ and $\ufdm$ at $t=1$, as well as the DICE scores $\mathrm{DICE}_m$. Figure~\ref{f:gbm} visualizes the predicted segmentations using BiLO and PINN for different values of $\wdat$.
For the PINNs, we observe that the DICE score based on $\unn$ is generally higher than that based on $\ufdm$, indicating a tendency to overfit the data while compromising the fidelity of the PDE solution. This behavior is reflected in the larger relative errors of $\unn$ and is visually apparent in Figure~\ref{f:gbm}, where the contours from $\unn$ track the noisy segmentation data more closely than those from $\ufdm$.

Reducing $\wdat$ helps mitigate this overfitting by regularizing the data fitting. In contrast, BiLO achieves both accurate PDE solutions and well-performing parameters without the need to tune $\wdat$, leading to better segmentations in this case. Interestingly, even when $\unn$ is not accurate in the PINN setting, the inferred parameters can still yield reasonable segmentation when evaluated using $\ufdm$, as evidenced by the corresponding DICE scores.

\begin{table}[!ht]
    \centering
    \caption{Results of the GBM inverse problem: Average DICE scores for the WT and TC regions based on predicted segmentations from the neural network solution $\unn$ and the numerical solution $\ufdm$, along with the relative mean squared error (MSE) of $\unn$ at $t=1$}
    \begin{tabular}{cccccc}
      \toprule 
      Methods & $\rm DICE_{NN}$ & $\rm DICE_{FDM}$& $\rm rel. MSE(\%)$\\ 
      \hline 
      BiLO       & 0.84  & 0.84 & 0.04 \\ 
      PINN($10^{-3}$) & 0.98  & 0.75 & 28 \\ 
      PINN($10^{-5}$) & 0.87  & 0.83 & 1 \\ 
      PINN($10^{-7}$) & 0.83  & 0.83 & 0.03 \\ 
      \bottomrule
      \end{tabular}
      \label{t:gbm}
\end{table}

\paragraph{Summary of Additional Results} 
\change{For each numerical experiments, the training details and some additional results are provided in \ref{ap:train}. 
In \ref{ap:hyper}, we demonstrate the robustness of BiLO with respect to the learning rate and the residual-gradient weight $\wdr$. 
These findings highlight the practical reliability of BiLO without requiring extensive hyperparameter tuning.
The computational costs, including seconds-per-step and peak memory usage, are provided in \ref{ap:cost}. 
While BiLO is more costly per step than PINN, overall it requires fewer steps to converge, leading to a lower total computational cost.}

\section{Conclusion}
\label{conclusion}
In this work, we presented a Bi-level Local Operator (BiLO) learning framework for solving PDE-constrained optimization problems.
We minimize the data loss with respect to the PDE parameters at the upper level, and learn the local solution operator of the PDE at the lower level.
The bi-level optimization problem is solved using simultaneous gradient descent, leading to an efficient algorithm.
Empirical results demonstrate more accurate parameter recovery and stronger fidelity to the underlying PDEs under sparse and noisy data,
compared with the soft PDE-constraint formulation, which faces the delicate trade-off between adhering to the PDE constraints and accurately fitting the data. 
\change{Future work includes a deeper theoretical analysis of the simultaneous gradient descent dynamics, reducing computational cost via Gaussian Process representations of the PDE solution and Gauss--Newton updates at the lower level \cite{chenSolvingLearningNonlinear2021,nelsenBilevelOptimizationLearning2025}, and extending BiLO to more complex settings such as full 3D tumor inverse problems \cite{zhangPersonalizedPredictionsGlioblastoma2025,balcerakIndividualizingGliomaRadiotherapy2024}.
}
\newpage

\section*{Code Availability}
The code for the numerical experiments is available at \url{https://github.com/Rayzhangzirui/BiLO}.
\section*{Acknowledgment}
R.Z.Z and J.S.L thank Babak Shahbaba for GPU resources. J.S.L acknowledges partial support from the National Science Foundation through grants DMS-2309800 and DMS-1763272 and the Simons Foundation (594598QN) for an NSF-Simons Center for Multiscale Cell Fate Research. C.E.M was partially supported by a NSF CAREER grant DMS-2339241.

\bibliographystyle{elsarticle-num}
\bibliography{bilopdeco.bib}

\newpage
\appendix




\section{Theoretical Analysis}
\label{ap:gd}
In the main text, we describe a simultaneous gradient descent algorithm for solving the bi-level optimization problem. In this section, we provide a theoretical justification for the algorithm under certain simplifying assumptions.

This appendix provides a rigorous bound on the error of the approximate hypergradient used in the BiLO framework, with results that apply to both PDE-constrained optimization setting in Part I and Bayesian inference in Part II \cite{zhangBiLOBilevelLocal2025}. 
While the proofs rely on highly idealized conditions, the resulting theory offers valuable insights into the theoretical behavior of BiLO. Numerical experiments in the main text further illustrate that BiLO remains effective even when some restrictive assumptions are relaxed.

We present our theoretical analysis through two theorems organized into distinct subsections to enhance clarity and accessibility. The first theorem (Theorem 1 in ~\ref{ap:consistency}) focuses on the consistency of the approximate gradient, showing that it is exact when the lower-level problem is solved to optimality.
It illustrates the key calculations in a simplified setting, specifically under the idealized condition of exact minimization at the lower level, and assuming a linear PDE operator with established stability properties. This special case offers a transparent view into why BiLO’s approximate gradient can be exact under ideal circumstances. 
The second theorem (Theorem 2, in \ref{ap:approximation}) extends this analysis into a more general and abstract setting, focusing on the approximation error introduced by the inexact minimization of the lower-level problem. Although Theorem 1 can technically be viewed as a special case or a corollary of Theorem 2, explicitly stating and proving it separately aids in highlighting the fundamental mechanism underlying BiLO and facilitates understanding of the broader and more general error analysis that follows.


\subsection{Consistency}
\label{ap:consistency}
\paragraph{Setup}
We consider the minimization problem
\[\min_\theta \norm{u-\hat{u}}\]
where $\hat{u}$ is the observation, and $u$ is the solution of the PDE
\begin{equation}
\begin{cases}
\bL u=f & \text{in }\Omega\\
u=g & \text{on }\partial\Omega,
\end{cases}
\end{equation}
where $\Omega$ is a connected, open and bounded subset of $\mathbb{R}^{d}$. $\bL$ denotes a second-order partial differential operator:
\begin{equation}
  \bL u = \sum_{i,j=1}^{d}a_{ij}\partial_{ij}u+\sum_{i=1}^{d}b_{i}\partial_{i}u+cu
  \label{apeq:elliptic}
\end{equation}
where the coefficients $a_{ij}$, $b_{i}$, $c$ are collectively denoted as $\theta$.
We denote $\bL_{\theta}$ as the Fréchet derivative of $\bL$ with respect to $\theta$, which is also a differential operator.

We define the \textbf{ideal optimal weight} as the exact minimizer of the following problem:
\[
W^*(\theta) = \arg\min\int_{\Omega}\left(\bL u(\theta,W)-f\right)^{2}d\bx
\]
That is, for all $\theta$
\begin{equation}
  \label{ustar}
  \bL u(\theta,W^*(\theta))=f.
\end{equation}
Denote $u^* = u(\theta, W^*)$ and define
\begin{equation}
    v:= d_\theta u(\theta, W^*(\theta)) = \nabla_{\theta}u^* + \nabla_{W}u^*\nabla_{\theta}W^*
\end{equation}

The exact hypergradient is given by
\begin{equation}
\gtrue(\theta) = \int_{\Omega}\left(u^*-\hat{u}\right)vd\bx
\end{equation}

In BiLO, the gradient of the upper level objective with respect to $\theta$ is given by
\begin{equation}
\gapprox(\wbar, \theta) = \int_{\Omega}\left(u(\wbar,\theta)-\hat{u}\right)\nabla_{\theta}u(\wbar,\theta) d\bx
\label{apeq:ga}
\end{equation}
where $\wbar$ is the minimizer of the lower level problem
\[
\wbar=\arg\min\int_{\Omega}\left(\bL u-f\right)^{2}+ \wreg \left(\bL_{\theta}u+\bL\nabla_{\theta}u\right)^{2} d\bx
\]

\begin{theorem}[Consistency of the approximate gradient]
\label{thm:consistency}
Assuming 
(i) The maximum principle holds for the PDE operator;
(ii) The parametrized local operator $u(\theta,W)$ satisfies the boundary condition for all $\theta$ and $W$;
(iii) The lower-level optimization is solved exactly.
then the approximate gradient of the upper level objective is exact.
\end{theorem}

\begin{proof}

Taking the total derivative of \eqref{ustar} with respect to $\theta$, we have 
\begin{equation}
    \label{dtheta_wstar}
 2   \bL_{\theta}u^*+\bL[\nabla_{\theta}u^* + \nabla_{W}u^*\nabla_{\theta}W^*]=\bL_{\theta}u^*+\bL v=0
\end{equation}
By assumption (iii), the residual loss and the residual-gradient loss vanish exactly at the solution $\ubar$:
\begin{equation}
    \label{lbaru}
    \bL \ubar = f
\end{equation}
\begin{equation}
    \label{dlbaru}
    \bL_{\theta}\ubar + \bL\nabla_{\theta}\ubar = 0
\end{equation}
By the uniqueness of the solution to the PDE, we have $u^* = \ubar$.
Subtract \eqref{dtheta_wstar} from \eqref{dlbaru}, we obtain
\begin{equation}
\bL[\nabla_\theta\ubar - v] = 0
\end{equation}
Since $u(\theta,W) = g$ on $\partial\Omega$ for all $W$ and $\theta$,
we have $\nabla_\theta\ubar - v=0$ on $\partial\Omega$.
If the maximum principle holds for the operator $\bL$ (for example, when $\bL$ uniformly elliptic and $c\geq0$ \cite{evansPartialDifferentialEquations2010}), then $v - \nabla_\theta\ubar = 0$.

The difference between the exact gradient and the approximate gradient, which we denote as $\Delta g$, is given by
\begin{equation}
\begin{aligned}
  \Delta g &= \gapprox(\wbar, \theta) - \gtrue(\theta) \\
  &= \int_{\Omega}\left(u^*-\hat{u}\right)\left(\nabla_\theta\ubar - v \right) d\bx  
\end{aligned}
\end{equation}
By the Cauchy-Schwarz inequality, we have 
\begin{equation}
  \norm{\Delta g} \leq \norm{u^*-\hat{u}}\norm{\nabla_\theta\ubar - v} = 0
\end{equation}
That is, the approximate gradient at $\wbar$ is exact. 
\end{proof}

These assumptions are more restrictive than the numerical experiments. 
For example, in the Fisher-KPP example, the PDE operator is nonlinear. 
A more general analysis is shown in the next section, where we bound the error of the approximate gradient by the lower-level optimization error.

\subsection{Approximation Error}
\label{ap:approximation}
\paragraph{Setup}
We consider the following PDE-constrained optimization problem:
\begin{equation}
  \begin{aligned}
    &\min_{\theta} \quad \cJ[u] \\
    &\textrm{s.t.} \quad \cF[u, \theta] = \mathbf{0}\\
  \end{aligned}
\end{equation}
on an open and bounded domain $\Omega \subset \mathbb{R}^d$. The PDE parameter $\theta \in \mathbb{R}^m$. 
The PDE residual operator $\mathcal{F}: H^1_0(\Omega) \times \mathbb{R}^m \to H^{-1}(\Omega)$.
The objective function $\cJ: H^1_0(\Omega) \to [0, \infty)$.
Denote the functional derivative of $\cJ$ by $\cJ'[\cdot]$.
We consider the PDE solution map, parameterized by weights $W \in \mathbb{R}^n$,
$u: \mathbb{R}^m \times \mathbb{R}^n \to H^1_0(\Omega)$. 
The residual as a function of $\theta$ and $W$ is defined as
\begin{equation}
  r(\theta,W) := \cF[u(\theta,W), \theta]
\end{equation}
Denote the (partial) Fréchet derivative of $\cF$ by $\cF_u$ and $\cF_\theta$. 
The residual-gradient is defined as
\begin{equation}
  \nabla_\theta r( \theta,W):= \cF_u(u, \theta)[\nabla_\theta u(\theta,W)] + \cF_\theta(u(\theta,W), \theta)
\end{equation}

Denote the linearized PDE operator at $u$ as
\[\bL_u[\cdot] := \cF_u(u, \theta)[\cdot]\]

The \textbf{ideal optimal weights} $W^*(\theta)$ satisfies the PDE for all $\theta$:
\begin{equation}
\label{wstarideal}
  \cF[u(\theta,W^*(\theta)), \theta]= \mathbf{0}
\end{equation}
This assumes that both the neural network’s intrinsic approximation error and the error due to using a finite set of collocation points are negligible.

Its practical \textbf{approximation} is denoted by $\wbar$, which is obtained by terminating the optimization once the lower-level loss is within a specified tolerance, 
$\llo(\theta, \wbar) \le \epsilon$. 
We also denote $u^* = u(\theta, W^*)$ the solution at the ideal optimal weights $W^*$.  $\hat{u} = u(\theta, \wbar)$ is the solution at the approximate weights $\wbar$.

For BilO, since the residual-gradient weight $\wdr<1$ is fixed, we may assume without loss of generality that both the residual loss and the residual-gradient loss are within a specified tolerance $\epsilon$:
\begin{equation}
    \norm{\cF( \ubar, \theta) }\leq \epsilon
\end{equation}
\begin{equation}
    \label{lowereps}
    \norm{ \bL_{\ubar}[\nabla_\theta \ubar] + F_\theta( \ubar, \theta)} \leq \epsilon
\end{equation}

Denote:
\begin{equation}
    v: = \nabla_\theta u(W^*, \theta) + \nabla_W u(W^*, \theta) \nabla_\theta W^*(\theta)
\end{equation}
which is the sensitivity of the ideal PDE operator $u^*$ with respect to $\theta$

The true hypergradient of the upper level objective $\cJ$ is given by
\begin{equation}
    \gtrue(\theta): = \nabla_\theta \cJ[u(\theta,W^*(\theta))] =  \cJ'[u^*]v
\end{equation}

The approximate gradient in BiLO is given by
\begin{equation}
    g_{a}(\theta) := \nabla_\theta \cJ[u(\theta, \wbar)]  = \cJ'[\ubar] \nabla_\theta \ubar
\end{equation}

We use $C$ to denote a generic constant that includes the stability constant, lipschitz constant, etc., which may vary from line to line.

\begin{theorem}[Approximate Gradient Error Bound]
\label{thm:hypergrad_err_bound}
Assuming 
(1) The PDE operator $\cF$ is Lipschitz continuous with respect to $u$ and $\theta$, 
(2) The PDE operator $\cF(\cdot, \theta)$ is stable; that is, if $\cF(u,\theta) = 0$ and $\norm{\cF(v, \theta)}\leq \varepsilon$, then $\norm{v-u}\leq C \varepsilon$ for some constant $C$.
(3) The linearized PDE operator $\bL_u$ is Lipschitz continuous with respect to $u$ and is stable,
(4) The objective functional $\cJ$ is Lipschitz continuous and has bounded a derivative.
Then
\[
\norm{g_{\rm a} - g_{\rm true}} = O(\epsilon)
\]

\end{theorem}

\begin{proof}

Taking the total derivative of \eqref{wstarideal} with respect to $\theta$, we have
\begin{equation}
    \label{residual_dtheta}
    \bL_{u^*}[v] + F_\theta(u^*, \theta) = 0
\end{equation}
Because $\norm{\cF( \ubar, \theta) }\leq \epsilon$ and $\cF(u^*, \theta) = 0$,
by the stability assumption on $\cF$, we have $\norm{\ubar - u^*}  = O(\epsilon)$.
With Lipschitz continuity of $F_\theta$, we have $\norm{\cF_\theta(\ubar,\theta) - F_\theta(u^*,\theta) }  = O(\epsilon)$.
From \eqref{lowereps} and the definition of $v$, we have:
\begin{equation}
    \norm{ \bL_{\ubar}[\nabla_\theta \ubar] - \bL_{u^*}[v]  + \cF_\theta( \ubar, \theta) - \cF_\theta(u^*, \theta)} \leq \epsilon
\end{equation}
Since $\norm{\cF_\theta(\ubar, \theta) - \cF_\theta(u^*, \theta)}$ is $O(\epsilon)$, by the triangle inequality, this implies:
\begin{equation}
    \norm{\bL_{\ubar}[\nabla_\theta\ubar] - \bL_{u^*}[v]} \leq \epsilon + \norm{\cF_\theta(\ubar, \theta) - \cF_\theta(u^*, \theta)} = O(\epsilon)
\end{equation}
By the Lipschitz continuity of $\bL_u$ with respect to $u$, we have
\begin{equation}
    \norm{\bL_{\ubar}[\nabla_\theta\ubar] -  \bL_{u^*}[\nabla_\theta\ubar]} \leq \norm{\bL_{\ubar}-\bL_{u^*}}\norm{\nabla_\theta\ubar} \leq C \norm{\ubar - u^*} = O(\epsilon)
\end{equation}
By \eqref{residual_dtheta}, we have
\begin{equation}
    \norm{\bL_{\ubar}[\nabla_\theta\ubar] - \bL_{u^*}[v]} = \norm{\bL_{\ubar}[\nabla_\theta\ubar] -  \bL_{u^*}[\nabla_\theta\ubar] + \bL_{u^*}[\nabla_\theta\ubar - v]}
\end{equation}
By the triangle inequality, we have
\begin{equation}
    \norm{\bL_{u^*}[\nabla_\theta\ubar-v]} \leq 
    \norm{ \bL_{\ubar}[\nabla_\theta\ubar] -  \bL_{u^*}[\nabla_\theta\ubar] }  
    + \norm{\bL_{\ubar}[\nabla_\theta\ubar] - \bL_{u^*}[v]} 
    = O(\epsilon)
\end{equation}
By the stability of the operator $\bL_{u^*}$,
\begin{equation}
    \norm{\nabla_\theta\ubar - v} = O(\epsilon)
\end{equation}
By the Lipschitz continuity of $\cJ'$, we have
\begin{equation}
    \norm{\cJ'[\ubar] - \cJ'[u^*]} \leq C \norm{\ubar - u^*} = O(\epsilon)
\end{equation}
By the boundedness of $\cJ'$ and $\norm{\nabla_\theta \ubar}$, we have the difference
\begin{equation}
    \begin{aligned}
        &\norm{g_{a} - g_{true}} \\
        =& \norm{\cJ'[\ubar] \nabla_\theta \ubar -\cJ'[u^*]v} \\
        \leq& \norm{\cJ'[\ubar] -\cJ'[u^*]} \, \norm{\nabla_\theta \ubar} + \norm{\cJ'[u^*]} \, \norm{\nabla_\theta \ubar - v} \\
        =& O(\epsilon)
    \end{aligned}
\end{equation}

In summary, the approximation error of the hypergradient is bounded by a constant multiple of $\epsilon$, where this constant depends on problem-specific parameters such as the smoothness and stability of the PDE operator $\cF$, the functional $\cJ$, and the parameterization of the solution $u$.
\end{proof}

\section{Comparison with BPN}
\label{ap:bpn}

BPN \cite{haoBilevelPhysicsInformedNeural2023} is motivated by the same concern about the trade-off between the data loss and the PDE loss in a penalty-like formulation in PINN. 
BPN follows the PINN framework: the solution of the PDE is represented by a neural network $u(x;W)$ ($\theta$ is not input to the network). The definition of data loss and the residual loss is the same as in PINN. However, in BPN, the residual loss is separate from the data loss, leading to the bilevel optimization problem
$$ 
\begin{aligned}
&\min_{\theta}  \ldatpinn(W^*(\theta)) \\
\text{s.t.} \quad& W^*(\theta) = \arg \min_W \lrespinn(W,\theta).
\end{aligned}
$$
The gradient of the data loss with respect to the PDE parameters is given by the chain rule
$$
\dv{\ldat}{\theta} = \dv{\ldat(W^*(\theta))}{W} \dv{W^*(\theta)}{\theta},$$
where the hypergradient is given by
$$\dv{W^*(\theta)}{\theta} = - \left[\frac{\partial^2 L_{res}}{\partial W \partial W^T}\right]^{-1} \cdot \frac{\partial^2 L_{res}}{\partial W \partial \theta^T}.$$
Broyden's method \cite{broydenClassMethodsSolving1965} is used to compute the hyper-gradient, which is based on the low-rank approximation of the inverse Hessian. 
In BPN, the bilevel optimization problem is solved iteratively. At each step, gradient descent is performed at the lower level for a fixed number of iterations, $N_f$. Following this, the hypergradient is computed using Broyden’s method, which requires $r$ iterations to approximate the inverse vector-Hessian product. This hypergradient is then used to perform a single step of gradient descent at the upper level.

The BiLO approach differs significantly. Instead of representing the PDE solution, BiLO represents the local PDE operator, leading to a different lower level problem that includes the residual-gradient loss.
This enables direct computation of gradients for $L{data}$ with respect to $\theta$, eliminating the need for specialized algorithms to approximate the hypergradient. This formulation also allows us to perform simultaneous gradient descent at the upper and lower levels, which is more efficient than the iterative approach in BPN. 
Our method is specialized for PDE-constrained optimization, leveraging the structure of the PDE constraint for efficiency (see the theorem in \ref{ap:gd}). In contrast, BPN adopts a more general bilevel optimization framework, which, while broadly applicable, does not fully exploit the unique characteristics of PDE problems.

\begin{figure}[!h]
  \centering
  \includegraphics[keepaspectratio,width=\linewidth]{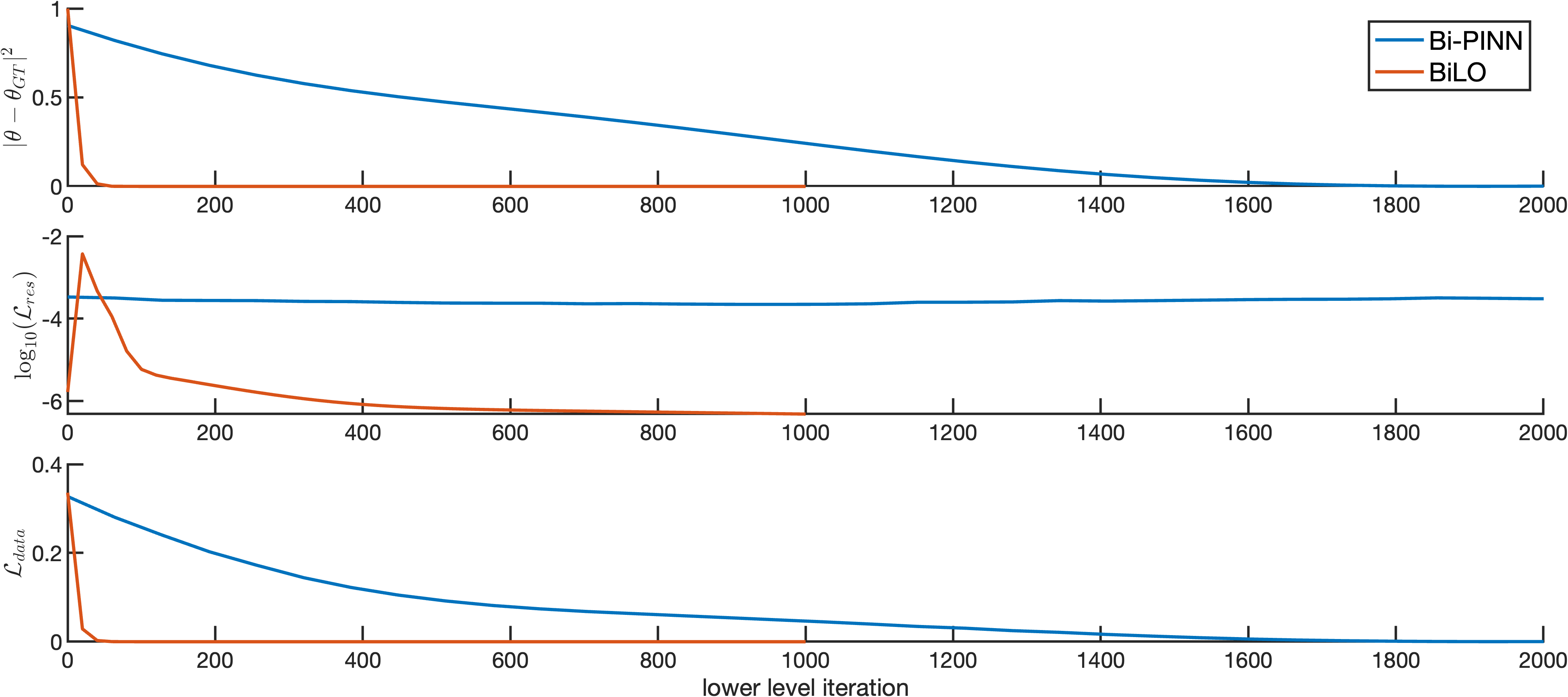}
  \caption{Comparison of BPN and BiLO methods. 
  x-axis is the number of lower level optimization steps.
  Top: Parameter error $\|\mathbf{\theta} - \mathbf{\theta}_{GT}\|^2$ versus iterations. 
  Middle: PDE loss $\log_{10}(\mathcal{L}_{res})$. 
  Bottom: Data loss $\mathcal{L}_{data}$.}
  \label{f:vsbipinn}
\end{figure}

To compare BiLO with BPN, we adopted the problem \eqref{eq:toy} and the setup from \cite{haoBilevelPhysicsInformedNeural2023}, using the same residual points (64), neural network architecture (4 hidden layers with 50 units), upper-level optimizer (Adam with learning rate 0.05), lower-level optimizer (Adam with learning rate 0.001), and initial guess ($\theta_0=0$,$\theta_1=1$). Both methods included 1000 pretraining steps to approximate the PDE solution at the initial parameters. In BPN, 64 lower iterations are performed for each upper iteration, with 32 Broyden iterations to compute the hypergradient. By contrast, BiLO performs simultaneous gradient descent at the upper and lower levels, where each iteration updates both levels concurrently.

\begin{equation}
  \begin{aligned}
    & \min_{\theta_0, \theta_1} J = \int_0^1 \left(y - x^2\right)^2 dx\\
    & \text{s.t.} \quad \frac{d^2y}{dx^2} = 2, \quad y(0) = \theta_0, \quad y(1) = \theta_1\\
  \end{aligned}
  \label{eq:toy}
\end{equation}
Figure~\ref{f:vsbipinn} presents the loss and the error of the PDE parameters for both methods versus the number of lower-level iterations. BiLO achieves a parameter error below 0.01 in fewer than 80 iterations and just 6.4 seconds, while BPN requires 27 upper iterations (1728 lower iterations) and 231 seconds to reach the same accuracy. While this highlights BiLO’s efficiency, we note that both methods may benefit from further hyperparameter tuning, and the comparison is made under the settings reported in \cite{haoBilevelPhysicsInformedNeural2023}.

\section{Training Details and Additional Results}
\label{ap:train}
In all the numerical experiments, we use the tanh activation function and 2 hidden layers, each with 128 neurons, for both PINN and BiLO.
The collocation points are evenly spaced as a grid in the domain.
For all the optimization problems, we use the Adam optimizer with learning rate 0.001 and run a fixed number of steps. 

\subsection{Fisher-KPP Equation} \label{ap:ss:fk}

\begin{figure}[!h]
  \centering
     \includegraphics[width=\linewidth,keepaspectratio]{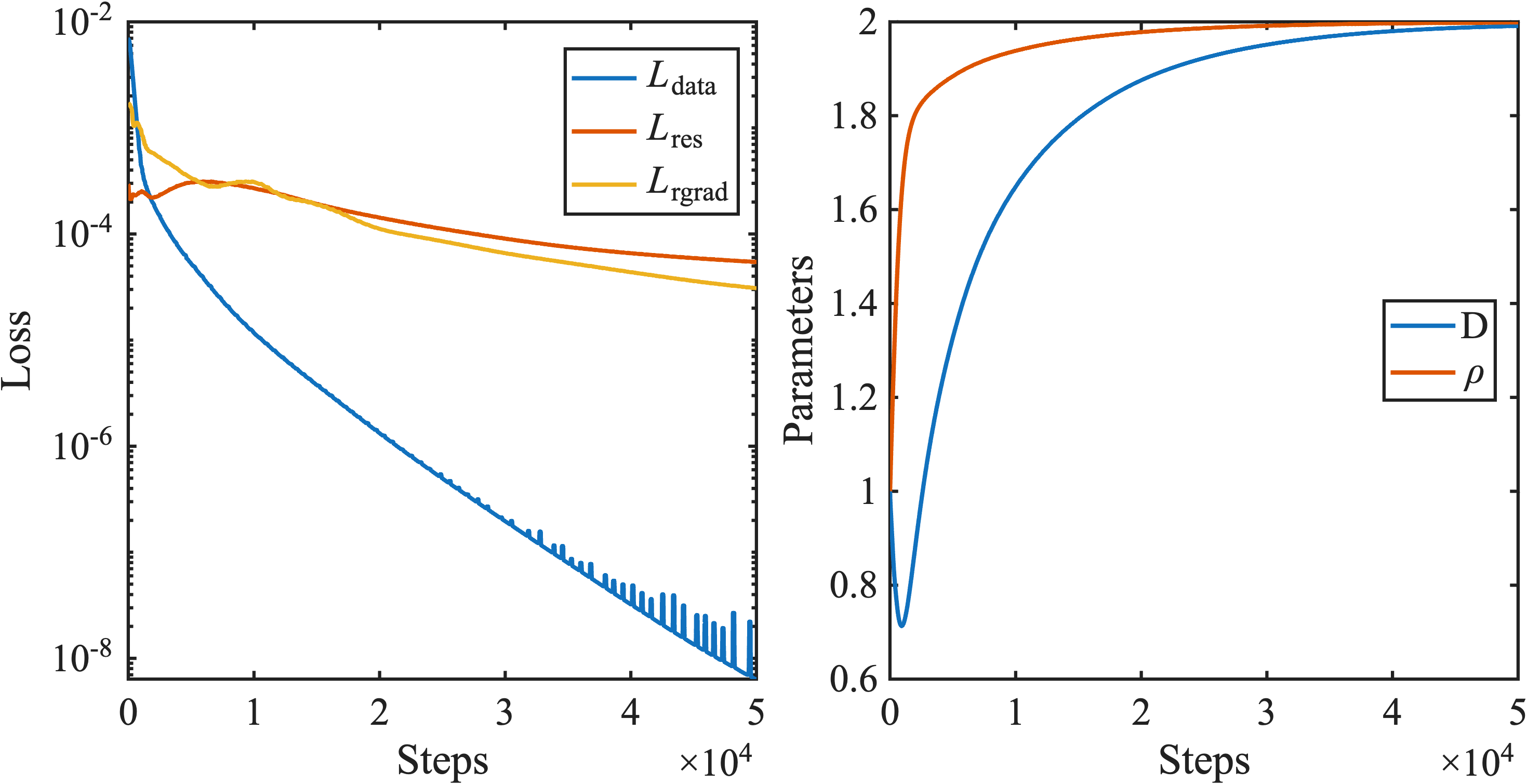}
   \caption{ Training history of the unweighted losses ( $\lres$, $\ldat$, and $\ldr$ ) and the PDE parameters ($D$ and $\rho$) during the fine-tuning stage for solving inverse problems using BiLO for the Fisher-KPP equation. 
   }
   \label{f:fklosstraj1}
 \end{figure}

Our local operator takes the form 
$$u(x,t,D,\rho;\wnn) = u(x,0) + \cN(x,t,D,\rho;\wnn)x(1-x)t$$ 
so that the initial condition and the boundary condition are satisfied.
Let $X_r$, $X_d$ be the spatial coordinates evenly spaced in $[0,1]$, and $T_r$ be temporal coordinates evenly spaced in $[0,1]$.
We set $\tdat = \tres = X_r \times T_r$ and $|X_r| = |T_r| = 51$. Both BiLO and PINN are pretrained with the initial guess for 10,000 steps, and fine-tuned for 50,000 steps.
In Fig.~\ref{f:fklosstraj1}, we show the training history of the inferred parameters and the inferred parameters during the fine-tuning stage for solving inverse problems using BiLO for the Fisher-KPP equation (Section \ref{ss:fk}, with no noise and $\wdr = 0.1$).

\subsubsection{Comparison with Neural Operators}\label{ss:fk:deeponet}
We use the following architecture for the DeepONet \cite{luLearningNonlinearOperators2021} in Section~\ref{ss:fk:deeponet}
\begin{equation}
  G_W(D,\rho,\vb{x}) = \sum_{i=1}^{k} b_k(D,\rho) t_k(\vb{x})
\end{equation}
where $b_k(D,\rho)$ is the k-th output of the ``branch net'', and $t_k(\vb{x})$ is the k-th output of the ``trunk net''. Both the branch net and the trunk net are parameterized by fully neural networks with 2 hidden layers, each with 128 neurons, so that the total number of parameters (46179) are comparable to the network used by BiLO (42051).
The weights of the DeepONet are denoted as $W$. 
A final transformation on the output $G_W$ is used to enforce the boundary condition.
We pre-train multiple DeepONets with 10,000 steps using each dataset. 

Given a pretrain dataset with collections of \{$D^{j}$, $\rho^{j}$\}
and their corresponding solutions $u^{{j}}$ for $j=1,\ldots,m$,
we first train the DeepONet with the following operator data loss:
\begin{equation}
  \min_W \sum_{j=1}^{m} \sum_{\bx\in \tdat}
  \left| G_{W}(D^{j},\rho^{j}, \bx) - u^{j}(\bx) \right|^2
\end{equation}
where $\tdat$ is the same as those used in the BiLO and PINN.
For the inverse problem, we fix the weights $W$ and treat the $D$ and $\rho$ as unknown variables.
We minimize the data loss:
\begin{equation}
  \min_{D,\rho} \frac{1}{|\tdat|} \sum_{\bx\in \tdat} \left| G_{W}(D,\rho, \bx) - \hu(\bx) \right|^2 
\end{equation}
where $\hu$ is the noisy data.

\subsection{Variable-Diffusion Coefficient Poisson Equation}
\label{ap:ss:varpoi}

The local operator takes the form of $u(x,z;W) = \cN_1(x,z;W)x(1-x)$ to enforce the boundary condition, where the fully connected neural network $\cN_1$ has 2 hidden layers, each with 128 neurons.
The unknown function is parameterized by $D(x;V) = \cN_2(x,V)x(1-x)+1$, where $\cN_2$ has 2 hidden layers, each with 64 neurons.
For pre-training, we set $|\tres| = |\treg| = |\tdat|= 101$, and train 10,000 steps.
For fine-tuning, we set $|\tres| = |\treg| = 101$ and $|\tdat|=51$, and train 10,000 steps. 

\subsubsection{Implementation of the Adjoint Methods}\label{ap:ss:varpoi-adjoint}
For the numerical example on learning the variable diffusion coefficient of the Poisson Equation, we implement the adjoint method following \cite{vogelComputationalMethodsInverse2002}.
The domain is discretized with uniformly spaced grid points:
$x_i  = hi$ for $i=0,\ldots,n,n+1$, where $h$ is the spacing of the grid points and $n$ is the number of intervals.
We use the finite element discretization with linear basis functions $\phi_i$.
Let $\bu$ be the nodal value of the solution $u$ at $x_i$ for $i=1,\ldots,n$ and similar for $\vb{D}$. We have $\bu_0 = \bu_{n+1} = 0$ and $\bD_0 = \bD_{n+1} = 1$. The stiffness matrix $A(\bD)$ is given by
\begin{equation}
  A(\bD)_{ij} = \frac{1}{2} 
  \begin{cases}
    \bD_{i-1} + 2\bD_i + D_{i+1}  & \text{if } i=j\\
    - (\bD_{i} + \bD_j) & \text{if } |i-j|=1\\
    0 & \text{otherwise}
  \end{cases}
\end{equation}
The load vector $\vb{f}$ is given by $\vb{f}_i = f(x_i)$.
Suppose the observed data is located at some subset of the grid points of size $m$.
Then $\hat{\vb u} = C\bu+\eta$, where $\eta$ is the noise, and $C \in \mathbb{R}^{n\times m}$ is the observation operator. 
After discretization, the minimization problem is
$$\min_{\mathbf{D}} || C\bu - \hat{\bu} ||_2^2 + \frac{w_{reg}}{2} \sum_{i=1}^{N} (\bD_{i+1}-\bD_i)^2 $$
$$\text{s.t } A(\mathbf{D}) \bu = \vb{f}$$

The gradient of the loss function with respect to the diffusion coefficient is given by
$$\vb{g}_i = \left<\frac{\partial A}{\partial D_i} \bu,\vb{z}\right> +{w_{reg}}\left( \bD_{i+1}-2\bD_{i}+\bD_{i-1} \right)$$
where $\vb{z}$ is the solution of the adjoint equation $A^T \vb{z} = C^T(C\bu-\hat{\bu})$.
Gradient descent with step size $0.1$ is used to update $D$, and is stopped when the norm of the gradient is less than $10^{-6}$.

\subsubsection{Comparison with DeepONet}\label{ap:ss:varpoi-deeponet}
\begin{figure}[!h]
  \centering
  \includegraphics[width=\linewidth,keepaspectratio]{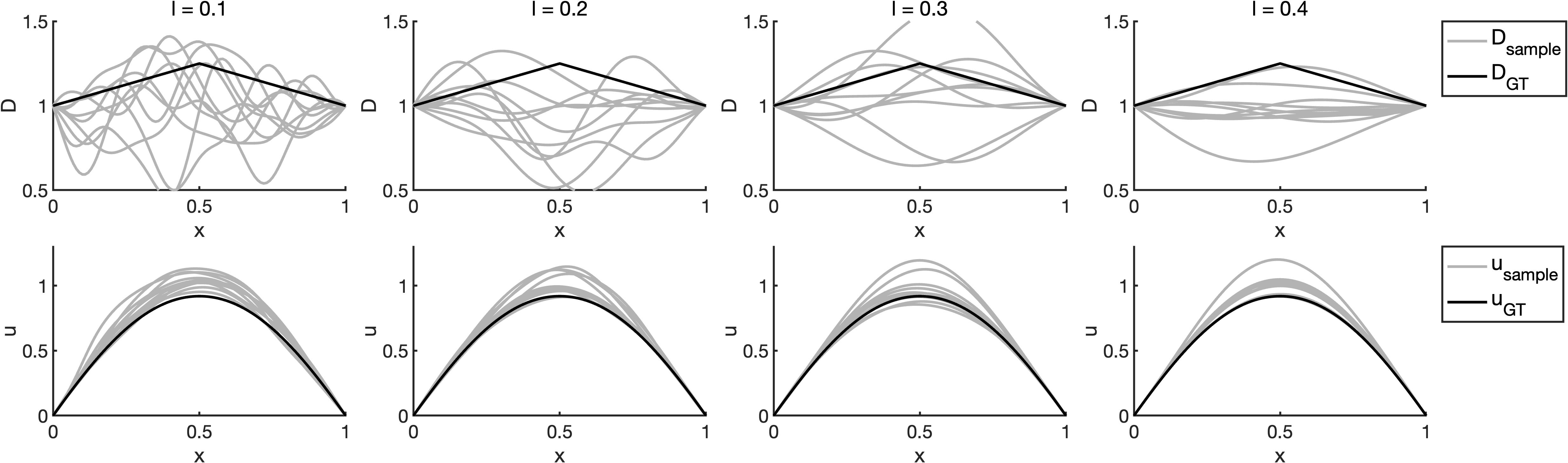}
  \caption{Samples (gray lines) of $D(x)$ with various length scale $l$ and their corresponding solutions. Black line is the ground truth $D$ and $u$}
  \label{f:sample}
\end{figure}

Figure \ref{f:sample} shows the samples of $D(x)$ with various length scale $l$ and their corresponding solutions $u$.

The DeepONet has the following architecture:
\begin{equation}
  G_W(\vb{D},\vb{x}) = \sum_{i=1}^{k} b_k(\vb{D}) t_k(\vb{x})
\end{equation}
where the vector $\vb{D}$ represent the values of $D(x)$ at the collocation points. A final transformation on the output $G_W$ is used to enforce the boundary condition.
In this experiment, both $D$ and $u$ are evaluated at 101 points in $[0,1]$.
Let $x_i$ be the collocation points in $[0,1]$ for $i=1,\ldots,N$. 
Let $\{D^{j}(x_i), u^{j}(x_i)\}$ be the samples of $D$ and the corresponding solutions $u$ at $x_i$ for $j=1,\ldots,m$. 
We denote $\vb{D}^{j}$ as the vector of $D^{j}(x_i)$ for $i=1,\ldots,N$.
In the pre-training step, we solve the following minimization problem
\begin{equation}
  \min_W \sum_{j=1}^{m} \sum_{i=1}^{N} \left| G_{W}(\vb{D}^{j}, x_i) - u^{j}(x_i) \right|^2
\end{equation}

For the inverse problem, we fix the weights $W$ and treat the $\vb{D}$ as an unknown variable.
We minimize the data loss and a finite difference discretization of the regularization term $|D(x)|^2$:
\begin{equation}
  \begin{aligned}
  &\min_{\vb{D}} \frac{1}{N} \sum_{i=1}^{N} \left| G_{W}(\vb{D}, x_i) - \hu(x_i) \right|^2 + \\
  &\wreg \sum_{i=0}^{N} \left| (\vb{D}_{i+1}-\vb{D}_{i})/h \right|^2 \\
  \end{aligned}
\end{equation}
where $h$ is the spacing of the collocation points, $\vb{D}_0 = \vb{D}_N = 1$. 
Here we work with the vector $\vb{D}$ for simplicity. Alternatively, we can represent $D(x)$ as a neural network as in PINN and BiLO experiments.

Table~\ref{t:varpoireg} presents the relative errors in the inferred diffusion coefficient $D$ and the solution $\unn$ across different methods and hyperparameter settings, under a fixed noise level $\sigma = 0.01$. For Neural Operator (NO), we vary the length scale $l = 0.2, 0.3, 0.4$ of the GRF in the pretrian dataset and regularization weight $\wreg = 10^{-5}, 10^{-4}, 10^{-3}$. The best performance is achieved with $l = 0.3$ and $\wreg = 10^{-4}$. 
For PINN, BiLO, and the Adjoint method, we consider $\wdat = 10, 100, 1000$, and evaluate each under $\wreg = 10^{-4}, 10^{-3}, 10^{-2}$. 
For PINN, we observe that both smaller $\wdat$ and larger $\wreg$ tend to promote smoother solutions, necessitating tuning of both hyperparameters. 
The combination $\wdat = 100$ and $\wreg = 10^{-3}$ gives the most accurate reconstruction among PINN variants. The BiLO method achieves the best overall performance with $\wreg = 10^{-3}$, outperforming both NO and PINN. 
For reference, the adjoint method solves the inverse problem numerically on a fine grid and is treated as the ground truth; hence, no relative error in $u_{\text{NN}}$ is reported. 

\begin{table}[!h]
\centering
\begin{tabular}{cccc}
\toprule
{Method} & ${\wreg}$ & {Rel Err. $D$} & {Rel Err. $u_{\text{NN}}$} \\
\midrule
\multirow{3}{*}{BiLO} 
    & $10^{-4}$ & $2.70\times10^{-2} \pm 6.74\times10^{-3}$ & $7.33\times10^{-4} \pm 3.00\times10^{-4}$ \\
    & $10^{-3}$ & $1.87\times10^{-2} \pm 8.25\times10^{-3}$ & $1.62\times10^{-3} \pm 9.87\times10^{-4}$ \\
    & $10^{-2}$ & $4.81\times10^{-2} \pm 2.13\times10^{-3}$ & $7.32\times10^{-4} \pm 7.29\times10^{-4}$ \\
\midrule
\multirow{3}{*}{PINN(10)} 
    & $10^{-4}$ & $2.94\times10^{-2} \pm 5.51\times10^{-3}$ & $4.76\times10^{-4} \pm 1.18\times10^{-4}$ \\
    & $10^{-3}$ & $3.20\times10^{-2} \pm 3.08\times10^{-3}$ & $7.25\times10^{-4} \pm 8.30\times10^{-4}$ \\
    & $10^{-2}$ & $4.84\times10^{-2} \pm 3.38\times10^{-3}$ & $5.20\times10^{-4} \pm 3.67\times10^{-4}$ \\
\midrule
\multirow{3}{*}{PINN(100)} 
    & $10^{-4}$ & $3.33\times10^{-2} \pm 1.32\times10^{-2}$ & $2.48\times10^{-4} \pm 1.89\times10^{-4}$ \\
    & $10^{-3}$ & $2.88\times10^{-2} \pm 8.10\times10^{-3}$ & $2.64\times10^{-4} \pm 2.13\times10^{-4}$ \\
    & $10^{-2}$ & $3.04\times10^{-2} \pm 7.48\times10^{-3}$ & $7.05\times10^{-4} \pm 4.80\times10^{-4}$ \\
\midrule
\multirow{3}{*}{PINN(1000)} 
    & $10^{-4}$ & $6.99\times10^{-2} \pm 3.66\times10^{-2}$ & $1.03\times10^{-3} \pm 4.51\times10^{-4}$ \\
    & $10^{-3}$ & $5.71\times10^{-2} \pm 2.57\times10^{-2}$ & $7.41\times10^{-4} \pm 5.14\times10^{-4}$ \\
    & $10^{-2}$ & $3.64\times10^{-2} \pm 1.32\times10^{-2}$ & $8.73\times10^{-4} \pm 5.71\times10^{-4}$ \\
\midrule
\multirow{3}{*}{Adjoint} 
    & $10^{-4}$ & $5.23\times10^{-2} \pm 1.36\times10^{-2}$ & - \\
    & $10^{-3}$ & $3.04\times10^{-2} \pm 9.77\times10^{-3}$ & - \\
    & $10^{-2}$ & $4.27\times10^{-2} \pm 4.35\times10^{-3}$ & - \\
\midrule
\multirow{3}{*}{NO (0.2)} 
    & $10^{-5}$ & $4.36\times10^{-2} \pm 1.01\times10^{-2}$ & $5.75\times10^{-3} \pm 1.05\times10^{-3}$ \\
    & $10^{-4}$ & $4.12\times10^{-2} \pm 8.69\times10^{-3}$ & $5.32\times10^{-3} \pm 9.58\times10^{-4}$ \\
    & $10^{-3}$ & $5.63\times10^{-2} \pm 2.89\times10^{-3}$ & $6.04\times10^{-3} \pm 4.12\times10^{-4}$ \\
\midrule
\multirow{3}{*}{NO (0.3)} 
    & $10^{-5}$ & $3.13\times10^{-2} \pm 1.05\times10^{-2}$ & $1.69\times10^{-2} \pm 1.20\times10^{-3}$ \\
    & $10^{-4}$ & $3.09\times10^{-2} \pm 9.68\times10^{-3}$ & $6.56\times10^{-3} \pm 1.59\times10^{-3}$ \\
    & $10^{-3}$ & $5.63\times10^{-2} \pm 3.27\times10^{-3}$ & $8.73\times10^{-3} \pm 9.74\times10^{-4}$ \\
\midrule
\multirow{3}{*}{NO (0.4)} 
    & $10^{-5}$ & $4.61\times10^{-2} \pm 8.54\times10^{-3}$ & $2.72\times10^{-2} \pm 4.34\times10^{-3}$ \\
    & $10^{-4}$ & $3.39\times10^{-2} \pm 9.10\times10^{-3}$ & $8.85\times10^{-3} \pm 3.69\times10^{-3}$ \\
    & $10^{-3}$ & $5.96\times10^{-2} \pm 3.77\times10^{-3}$ & $1.28\times10^{-2} \pm 1.73\times10^{-3}$ \\
\bottomrule
\end{tabular}
\caption{Relative errors in the inferred diffusion coefficient $D$ and neural network solution $\unn$ for different methods under noise level $\sigma = 0.01$. We vary the regularization strength $\wreg$ across all methods. For NO, we additionally vary the GRF length scale $l$ used in the pretraining data. For PINN, we vary the number of training data $\wdat = 10, 100, 1000$. The adjoint method serves as the reference solution and does not report error in $\unn$.}
\label{t:varpoireg}
\end{table}

\subsection{Infer the Initial Condition of a Heat Equation}\label{ap:ss:heat}

We can represent the unknown function $f(x;V) = s(\cN(x;V))x(1-x)$, where $N_f$ is a fully connected neural network with 2 hidden layers and width 64, and $s$ is the softplus activation function (i.e., $s(x) = \log(1+\exp(x))$).
The transformation ensures that the initial condition satisfies the boundary condition and is non-negative.
For BiLO, the neural network is represented as $u(x,t,z) = N_u(x,t,z;W)x(1-x)t + z$, where $N_u$ is a fully connected neural network with 2 hidden layers and width 128.
For the PINN, we have $u(x,t;\wnn,\wfcn)= N_u(x,t;\wnn)x(1-x)t + f(x;\wfcn)$.
These transformations ensure that the networks satisfy the boundary and initial conditions.

Let $X_r$, $X_d$ be spatial coordinates evenly spaced in $[0,1]$
and $T_r$ be temporal coordinates evenly spaced in $[0,1]$ (both including the boundary).
We set $\tres = X_r \times T_r$ and $|X_r| = |T_r| = 51$. That is, the residual collocation points is a uniform grid in space and time.
We set $\tdat = X_d \times \{1\}$ and $|X_d| = 11$. That is, the data collocation points is a uniform grid in space at the final time $t=1$.
We set the collocation point for the regularization loss of the unknown function $\treg$ to be 101 evenly spaced points in the spatial domain.
To evaluate the performance of the inferred initial condition $f$,
we use the relative $L^2$ norm of inferred initial condition and the ground truth initial condition,
which are computed using 1001 evenly spaced points in the spatial domain.

\subsection{Elliptic Equation with Singular Forcing}\label{ap:ss:pointproc}
For BiLO, the neural network is represented as 
$$u(x,\phi,\mu, \lambda;W) = N_u(x,\phi,\mu,\lambda;W)x(1-x),$$ 
where $N_u$ is a fully connected neural network with 2 hidden layers and width 128.
The FNO has 4 layer, 32 modes, and 32 channels. The FNO has 82785 parameters in total, the BiLO have 82691 parameters.

\subsection{1D Burgers' Equation}\label{ap:ss:burger}

\begin{figure}[!h]
  \centering
  \includegraphics[keepaspectratio,width=\linewidth]{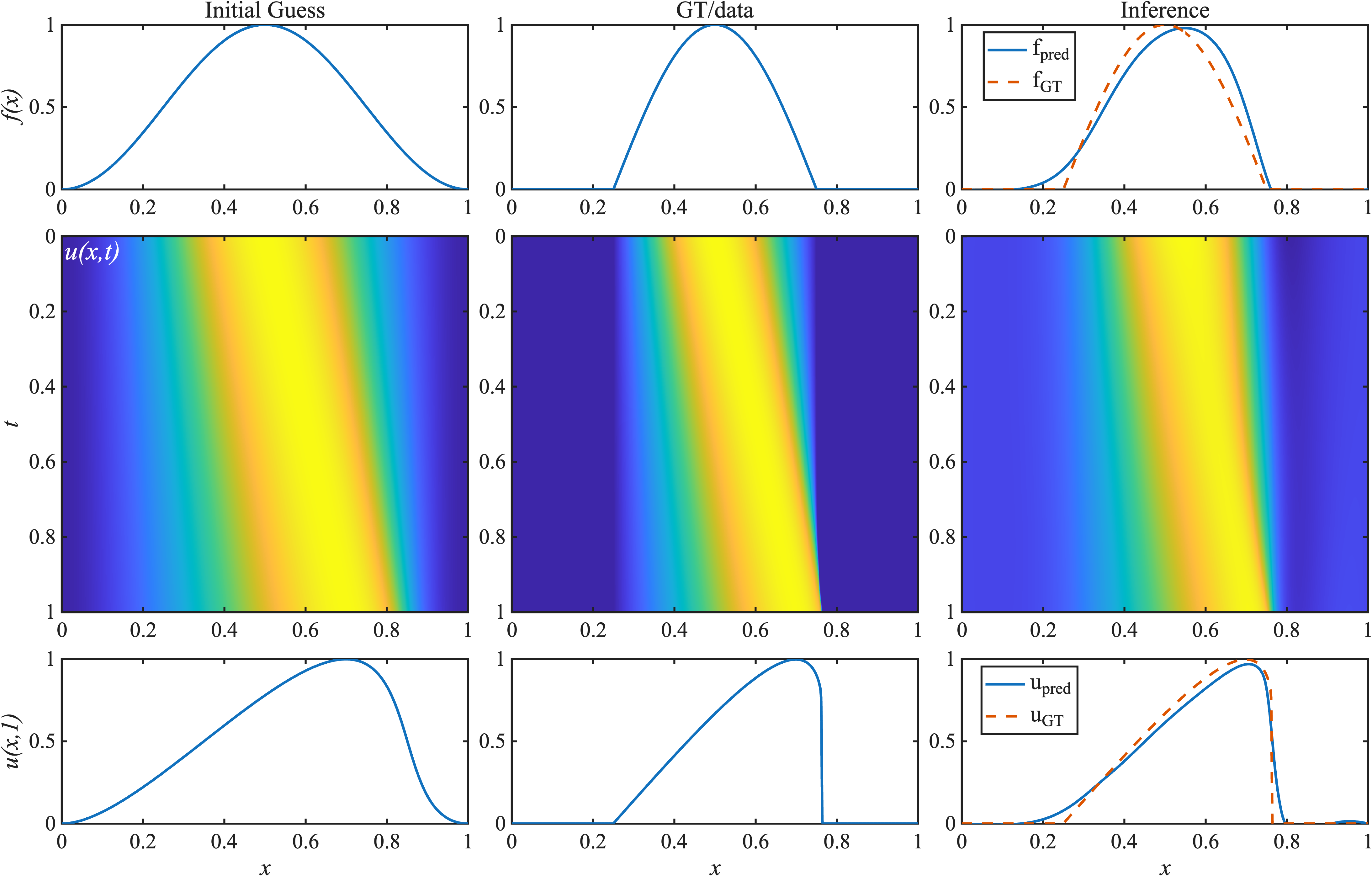}
  \caption{Example 2 of inferring the initial condition of the Burgers' equation.
  The initial guess is used to pre-train the network. The solution at $t=1$ of the GT is the data for inference. First column: initial guess, second column: ground truth, third column: inferred initial condition. First row: initial condition, second row: solution $u(x,t)$, third row: solution $u(x,1)$.
  }
  \label{f:burger2}
\end{figure}

Figure \ref{f:burger2} shows an additional example of the experiment in Section~\ref{ss:burger} in the main text.

\subsection{2D Darcy Flow}\label{ap:ss:2dpoisson}

The setup of this experiment is similar to the steady state Darcy flow inverse problem in \cite{liPhysicsInformedNeuralOperator2024}.
We pretrain the BiLO with $A_0(\vb{x})$ and it's corresponding solution $u_0(\vb{x})$ for 10,000 steps. And we fine-tune the BiLO for 5,000 steps using $u_{\rm GT}(\vb{x})$ to infer $A_{\rm GT}$. 

\begin{figure}[!h]
  \centering
  \includegraphics[keepaspectratio,width=\linewidth]{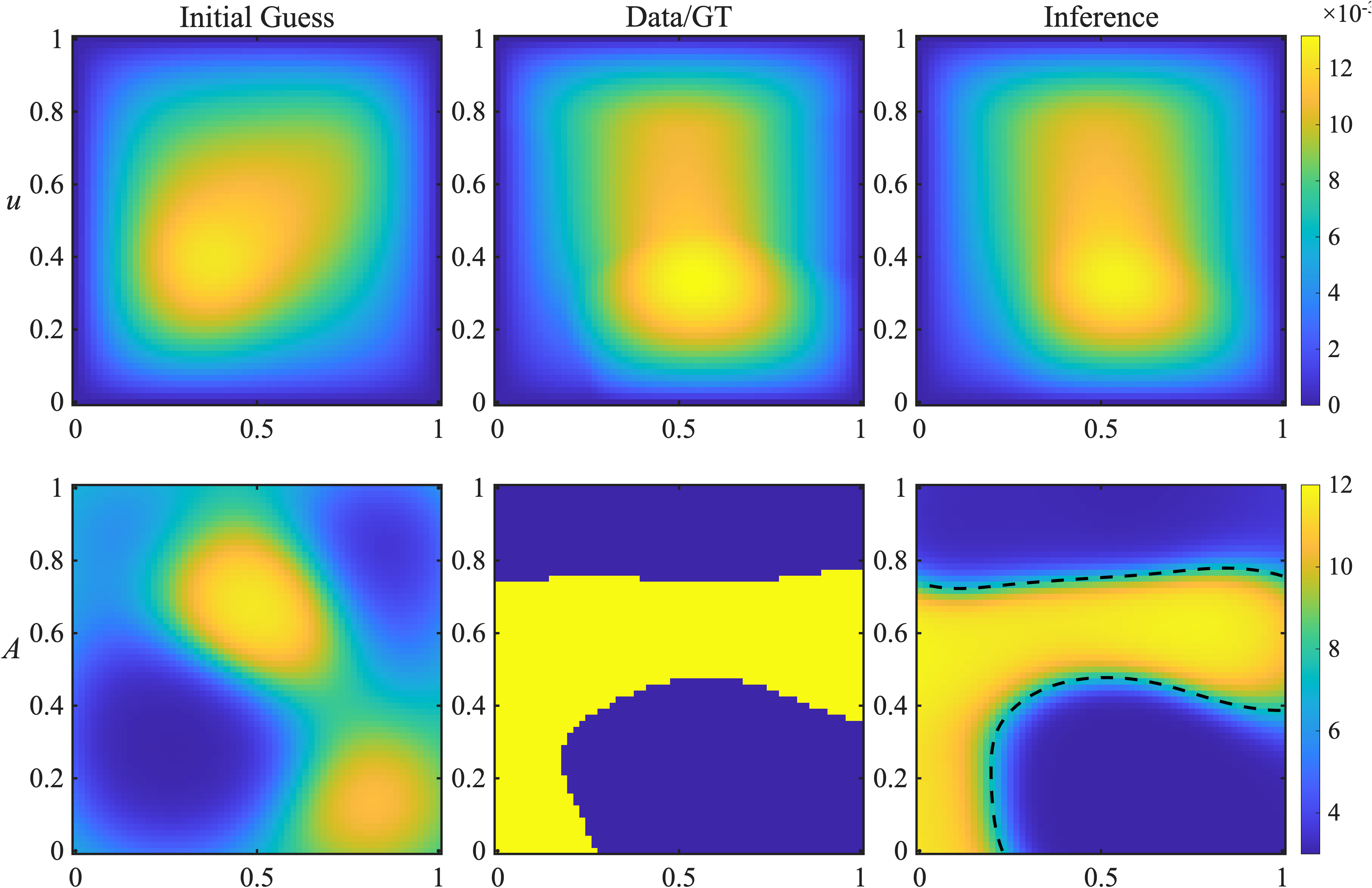}
  \caption{Example 2 of inferring the variable diffusion coefficient. The relative l2 error of $\unn$ against $\ugt$ is 1.7\%. 
  The thresholded (at the dashed line) inferred diffusion coefficient has classification accuracy of 96.7\%
  }
  \label{f:darcy2}
\end{figure}

The unknown function is represented as $A(\bx;V) = s(\cN(\bx;V))\times9+3$, where $N_f$ is a fully connected neural network with 2 hidden layers and width 64, and $s$ is the logistic function.
The transformation is a smoothed approximation of the piece-wise constant function.
For BiLO, the neural network is represented as $u(\bx,z) = N_u(\bx,z;W)\bx_1(1-\bx_1)\bx_2(1-\bx_2)$, where $N_u$ is a fully connected neural network with 2 hidden layers and width 128, and $z$ is our auxiliary variable such that $z = A(\bx;V)$.
Fig.~\ref{f:darcy2} shows an additional example of the experiment.

\section{Sensitivity to Hyperparameters} \label{ap:hyper}

We evaluate the sensitivity of BiLO to various hyperparameter choices in the noise-free FKPP inverse problem and compare it with PINN using the best-performing $\wdat = 10$. 

\begin{figure}[!h]
  \centering
  \includegraphics[width=\linewidth,keepaspectratio]{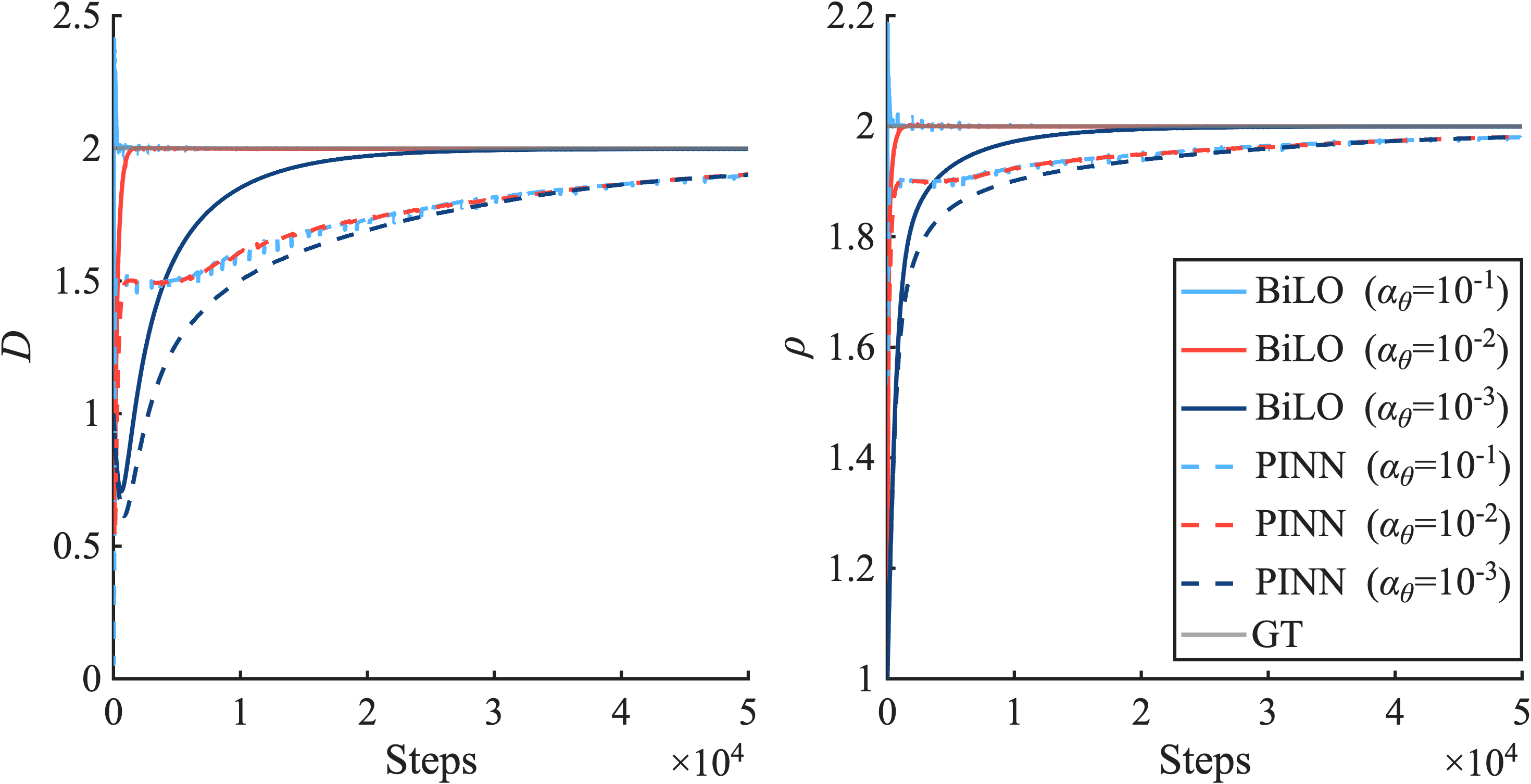}
  \caption{Trajectory of PDE parameters $D$ and $\rho$ with different $\alpha_\theta$. BiLO is robust to the choice of $\alpha_\theta$ and consistently outperforms.
  }
  \label{f:fklr}
\end{figure}

In both BiLO and PINN frameworks, it is possible to specify distinct learning rates for the PDE parameters $\theta$ and the neural network weights $W$, denoted as $\alpha_\theta$ and $\alpha_W$, respectively, as in equation~\eqref{eq:bilogd} and \eqref{eq:pinngd}.
In our experiments, we fixed $\alpha_W = 10^{-3}$, which is a common choice for training neural networks, and varied the learning rate $\alpha_\theta$ for both methods. 
As shown in Figure~\ref{f:fklr}, BiLO consistently achieves more accurate parameter recovery than PINN for any fixed choice of $\alpha_\theta$. Moreover, BiLO remains robust across a wide range of $\alpha_\theta$ values, including large values up to 0.1, and converges significantly faster without sacrificing accuracy. 
We attribute BiLO’s superior performance to its more accurate descent direction for $\theta$. 
Although we use $\alpha_\theta = 10^{-3}$ throughout the main text for consistent comparison, even better performance could likely be obtained through cross-validation over $\alpha_\theta$.

\begin{figure}[!h]
  \centering
     \includegraphics[width=\linewidth,keepaspectratio]{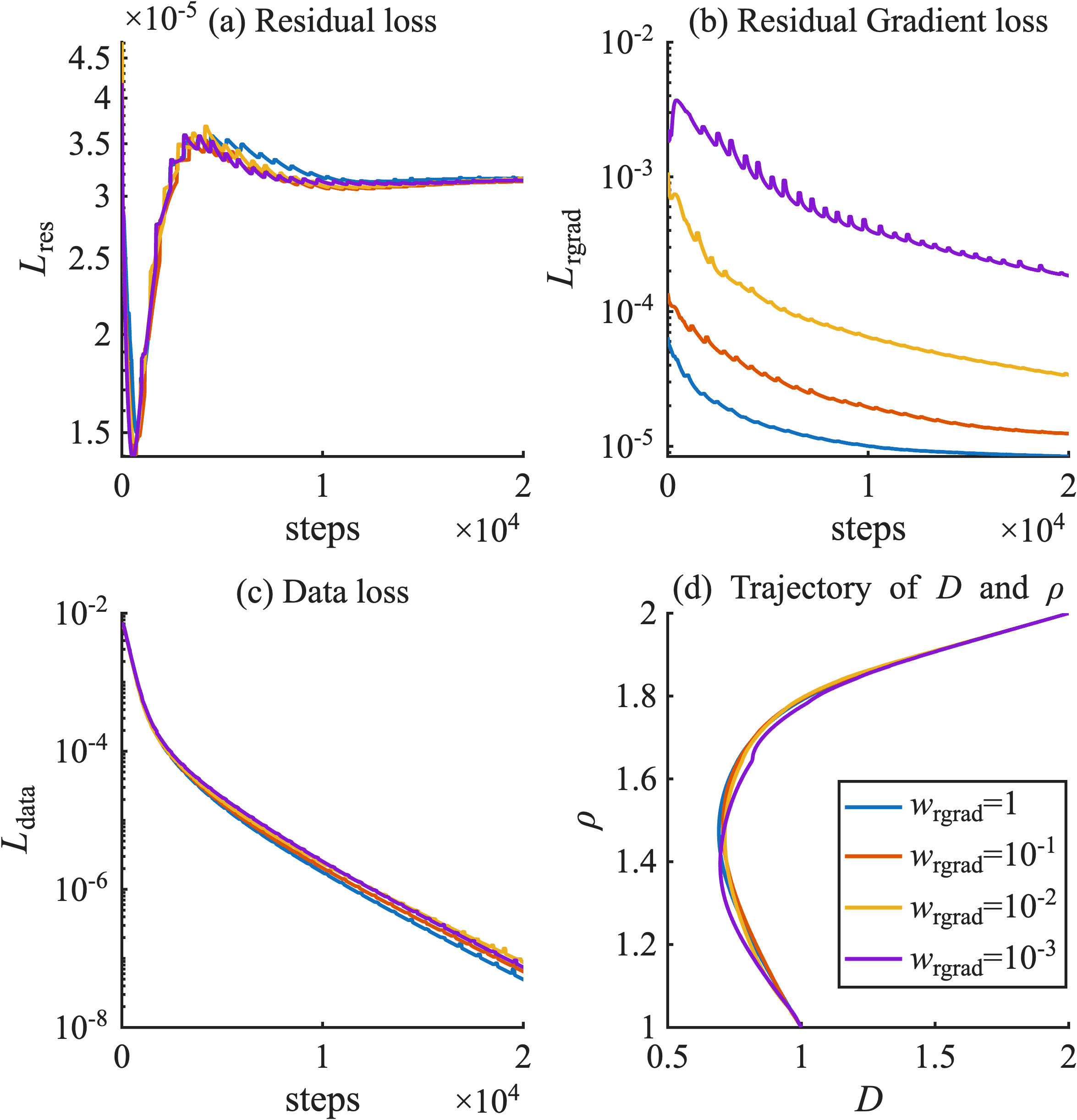}
   \caption{History of the losses—(a) $\lres$, (b) $\ldat$, and (c) $\ldr$—and (d) the trajectories of the parameters $D$ and $\rho$ for different values of the residual-gradient weight $\wdr$. 
   The initial guess of $(D,\rho)$ is $(1,1)$, and the ground truth is $(2,2)$.
   }
   \label{f:fkrgrad}
 \end{figure}

Figure~\ref{f:fkrgrad} illustrates that BiLO is robust with respect to the weight of the residual-gradient loss $\wdr$. 
Across different values of $\wdr$, the trajectories of the PDE parameters remain similar.
As $\wdr$ increases, the residual-gradient loss $\ldr$ decreases, but the residual loss $\lres$ remains small, confirming that the learned parameters still accurately solve the PDE. 
This suggests that $\ldr$ and $\lres$ exhibit a tradeoff distinct from that between $\ldat$ and $\lres$: while minimizing $\ldat$ alone may lead to functions that do not satisfy the PDE, both $\ldr$ and $\lres$ can, in principle, be minimized simultaneously if the local operator can be sufficiently approximated. 

\section{Computational Cost}\label{ap:cost}

Compared with PINN, BiLO involve computing a higher order derivative term in the residual-gradient loss. This increases the memory cost and computation time per step. In Table.~\ref{t:cost}, we show the seconds-per-step and the maximum memory allocation of 1 run of BiLO and PINN for the various problems. The seconds per step is computed by total training time divided by the number of steps. The maximum memory allocation is the peak memory usage during the training. For for all the experiments, we use Quadro RTX 8000 GPU. 
We note that the measured seconds-per-step is not subject to rigorous control as the GPU is shared with other users and many runs are performed simultaneously.

\begin{figure}[!h]
  \centering
     \includegraphics[width=\linewidth,keepaspectratio]{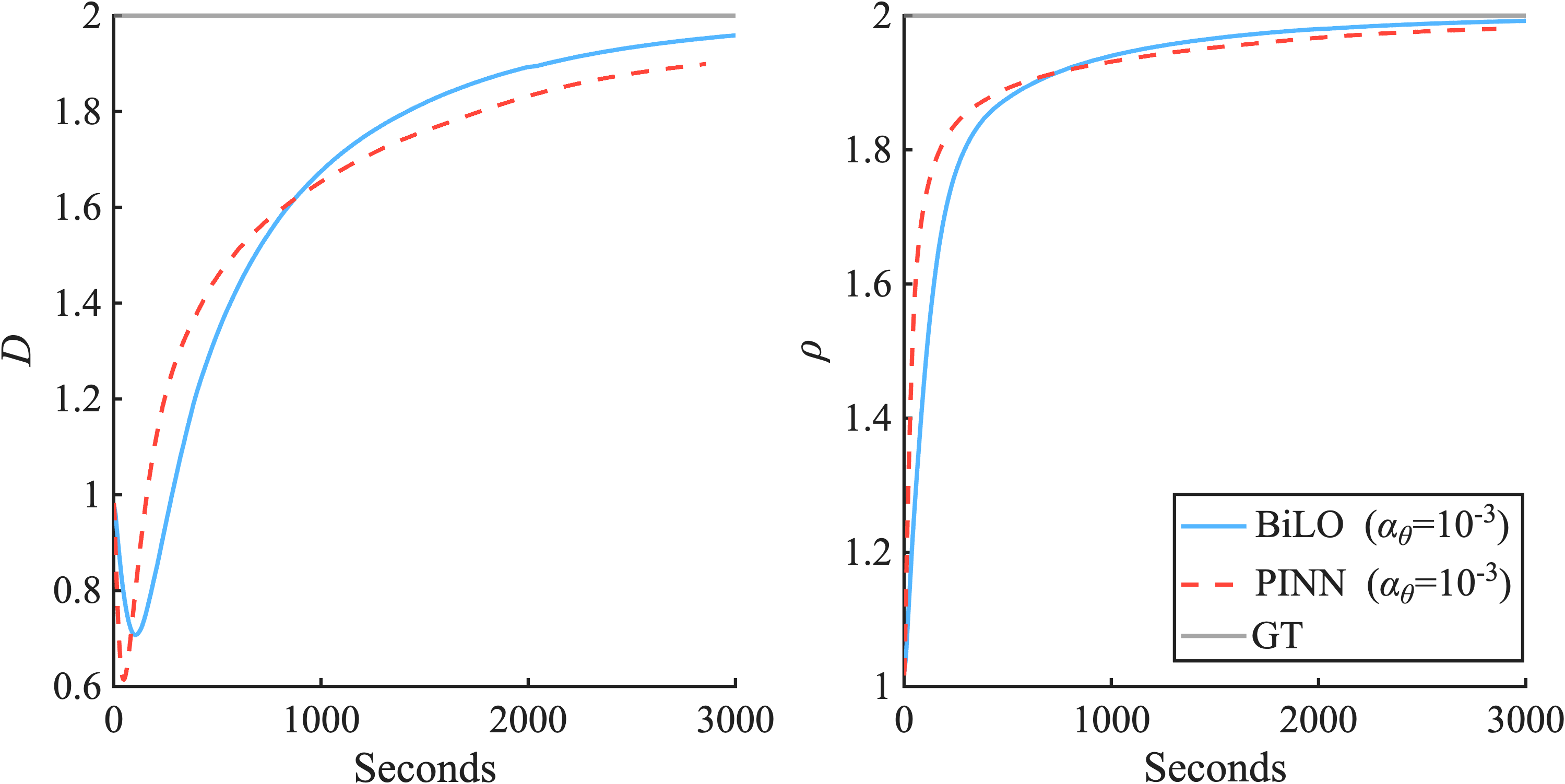}
   \caption{Trajectory of PDE parameters $D$ and $\rho$ during the fine-tuning stage for solving inverse problems using BiLO. The left panel shows the trajectory of $D$ and the right panel shows the trajectory of $\rho$. The x axis is the wall time in seconds. While BiLO takes more time per step than PINN, it converges faster to the optimal parameters. 
   }
   \label{f:fklosstraj}
 \end{figure}

While BiLO requires more computation per step, it achieves higher accuracy in less total time, as shown in Fig.~\ref{f:fklosstraj}, where we plot the trajectory of the PDE parameters with respect to wall time in seconds. We attribute this to the more accurate descent direction of the PDE parameters, which leads to faster convergence.

\begin{table}[!h]
  \centering
  \begin{tabular}{llccc}
  \toprule
  Problem & Metric & BiLO & PINN & Ratio \\
  \midrule
  \multirow{2}{*}{Fisher-KPP} 
    & sec/step & 0.11 & 0.06 & 1.69 \\
    & max-mem    & 200 & 65.2 & 3.07 \\
  \midrule
  \multirow{2}{*}{Singular}
    & sec/step & 0.21 & 0.11 & 2.00 \\
    & max-mem    & 67.4 & 44.9 & 1.50 \\
  \midrule
  \multirow{2}{*}{Poisson}
    & sec/step & 0.11 & 0.09 & 1.15 \\
    & max-mem    & 23.2 & 20.2 & 1.15 \\
  \midrule
  \multirow{2}{*}{Burgers}
    & sec/step & 0.17 & 0.08 & 2.24 \\
    & max-mem    & 122 & 73.9 & 1.65 \\
  \midrule
  \multirow{2}{*}{Heat}
    & sec/step & 0.11 & 0.07 & 1.57 \\
    & max-mem    & 211 & 109 & 1.93 \\
  \midrule
  \multirow{2}{*}{Darcy}
    & sec/step & 0.13 & 0.09 & 1.55 \\
    & max-mem    & 418 & 152 & 2.75 \\
  \midrule
  \multirow{2}{*}{GBM}
    & sec/step & 1.38 & 0.43 & 3.19 \\
    & max-mem    & 40313 & 10376 & 3.89 \\
  \bottomrule
  \end{tabular}
  \caption{Comparison of BiLO and PINN in terms of wall time per training step (in seconds) and maximum memory usage (in MB) across various PDE problems. Ratio is computed as BiLO / PINN. We note that the measured seconds-per-step is not subject to rigorous control as the GPU  (Quadro RTX 8000) is shared.}
  \label{t:cost}
\end{table}


\end{document}